\PassOptionsToPackage{table}{xcolor}
\documentclass[11pt]{article}

\usepackage[preprint]{acl}

\usepackage{times}
\usepackage{latexsym}

\usepackage[T1]{fontenc}

\usepackage[utf8]{inputenc}

\usepackage{microtype}

\usepackage{graphicx}
\usepackage{amsmath}
\usepackage{booktabs}
\usepackage{multirow}
\usepackage{xcolor}
\usepackage{placeins}
\usepackage{stfloats}
\usepackage{algorithm}
\usepackage{algpseudocode}
\definecolor{modelrow}{RGB}{242,242,242}
\definecolor{bestrow}{RGB}{255,247,235}

\title{MemGuard: Persisting Verifier Signals for LLM-Agent Memory Governance}

\author{
Haoyu Wang$^{1,*}$ \quad Guangyuan Dong$^{2,*,\dagger}$ \quad He Liang$^{3}$ \quad Zijing Zhang$^{4}$ \\
Jiachen Luo$^{5,6}$ \quad Chuang Liu$^{7}$ \quad Chao Xue$^{8}$ \quad Hao Tang$^{4,\dagger}$ \\
\normalfont\small $^1$Nankai University \quad $^2$National University of Singapore \\
\normalfont\small $^3$Shanghai Institute of Optics and Fine Mechanics, Chinese Academy of Sciences \\
\normalfont\small $^4$Peking University \quad $^5$Technical University of Munich \\
\normalfont\small $^6$Queen Mary University of London \quad $^7$Wuhan University \quad $^8$University of New South Wales \\
\normalfont\small $^*$Equal contribution. \quad $^\dagger$Corresponding authors.
}

\begin{document}
\maketitle
\begin{abstract}
LLM agents are moving from single-prompt use to long task streams in which reusable memory becomes a core capability for terminal, software-engineering, and web tasks. Such memory is useful only when stored experience remains reliable across hundreds of interactions, but two failure modes break that assumption in practice. The first is \emph{unreliable admission}: failed trajectories, accidental successes, and misleading observations enter memory because they appear relevant, then mislead later decisions. The second is \emph{memory drift}: long-running banks accumulate duplicate, stale, and conflicting records that retrieval alone cannot repair. MemGuard's key distinction is to treat verifier output not as a one-shot filter, but as persistent lifecycle metadata. It converts multi-criteria score-token verification into reward, confidence, label, and uncertainty descriptors that are attached to every candidate before activation and reused during retrieval, conflict resolution, summarization, and archival. We evaluate MemGuard on Terminal-Bench 2.0, SWE-Bench Verified, WebArena, and Mind2Web across four backbones, comparing against four memory baselines plus a verifier-only control under matched runtime budgets. Averaged over five seeds, MemGuard achieves the best success metric and lowest average steps in all 16 backbone--benchmark settings, improving over ReasoningBank, the strongest prior baseline among the memory methods we evaluate, with a largest gain of 7.9 success-rate points on WebArena, 5.6 step-success-rate points on Mind2Web, and 2.4--3.5 points on terminal and software-engineering benchmarks. Code is available at \url{https://github.com/whyyyyy123/MemGuard}.
\end{abstract}

\section{Introduction}

\begin{figure}[!t]
\centering
\includegraphics[page=1, width=\columnwidth]{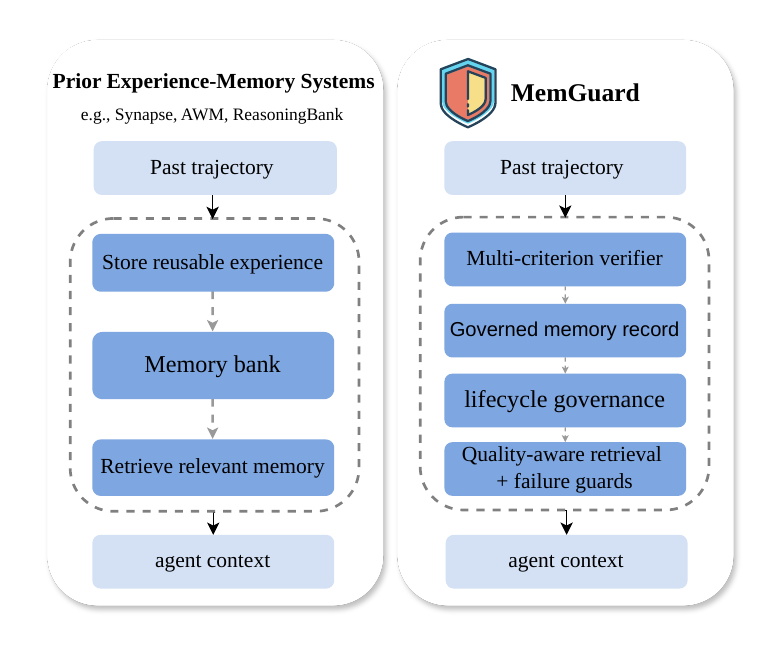}
\caption{Comparison between prior experience-memory systems such as Synapse \citep{zheng2024synapse}, AWM \citep{wang2024agentworkflowmemory}, and ReasoningBank \citep{ouyang2025reasoningbank} and MemGuard. MemGuard persists verifier-derived descriptors to govern memory admission, lifecycle updates, quality-aware retrieval, and failure-guard injection.}
\label{fig:memguard-pipeline}
\end{figure}

LLM agents are moving from single-prompt use to extended task sequences: software-engineering agents inspect repositories, edit code, and run tests; web agents navigate dynamic sites and execute multi-step actions; terminal agents plan, execute, and recover at the command line. In these settings, a single context window cannot preserve all useful prior experience, so production deployments increasingly depend on agent-memory systems that store trajectories, reflections, workflows, or reasoning hints for later reuse. The quality of this memory directly determines whether agents become more reliable as they accumulate experience or whether their behavior degrades unpredictably over time.

Two failure modes prevent agent memory from being reliable by default. First, \emph{unreliable admission}: interactive tasks routinely produce failed trajectories, accidental successes, misleading web observations, invalid patches, and outdated commands, any of which is distilled into memory whenever it looks relevant---for instance, a web agent that learns ``edit the first row after filtering an admin table'' will edit the wrong row whenever a later table is sorted differently. Second, \emph{memory drift}: once a bank grows across many tasks, it accumulates duplicate, conflicting, stale, and over-generalized records that remain retrievable long after their assumptions stop holding---for instance, a tool-usage hint from an older API version stays in the top-$k$ pool and is re-injected for every superficially similar query.

These failure modes call for two design principles that target the gaps left by retrieval alone. Before storing a memory, the system must judge whether a trajectory is successful, supported by evidence, executable, and generalizable. It must also store failed experience as a constraint against repeated mistakes rather than as an action recipe. After storage, memory cannot remain a static retrieval index: uncertain records must stay provisional, weak records must lose weight, duplicates must merge, conflicts must resolve, and outdated entries must be summarized or archived under a finite active-memory budget. Thus the central question is not whether an LLM verifier can screen memories once, but whether verifier evidence can persist as a control signal throughout the memory lifecycle.

We instantiate these principles in \textbf{MemGuard}, a verifier-guided memory governance framework whose contributions are: \textbf{(i) Verifier-guided admission.} MemGuard decomposes trajectory verification into multiple criteria, estimates reward from score-token distributions and repeated views, and attaches a persistent descriptor $d_m=(R_m,c_m,\ell_m,\nu_m)$ to every candidate before activation, so that low-quality experience is rejected, kept provisional, or routed to a failure-guard pool. \textbf{(ii) Structured memory governance.} Each record carries lifecycle state, quality, confidence, usage statistics, and conflict links, and $d_m$ continues to gate retrieval, conflict resolution, summarization, and archival throughout the record's lifetime---turning the memory bank into a lifecycle-managed object rather than an append-only index. \textbf{(iii) Comprehensive evaluation.} We evaluate MemGuard on Terminal-Bench 2.0, SWE-Bench Verified, WebArena, and Mind2Web across four backbones (Qwen-3.5-Flash, Qwen-3.5-Plus, Gemini-3-Flash, and Gemini-3.1-Pro), against four memory baselines (No Memory, Synapse, AWM, ReasoningBank) plus a verifier-only control, under matched task order, step budget, retrieval budget, and decoding settings; MemGuard achieves the best success metric in every backbone--benchmark setting, with up to 7.9 SR-point gains on WebArena and 5.6 SSR-point gains on Mind2Web over the strongest prior baseline among the memory methods we evaluate.

The verifier-only control isolates governance metadata: it improves over ReasoningBank in most cells by accepting better candidates, but underperforms MemGuard in every benchmark--backbone setting, showing that the gain comes from \emph{persisting} verifier signals rather than one-time filtering.

\begin{figure*}[t]
\centering
\includegraphics[page=1, width=\textwidth]{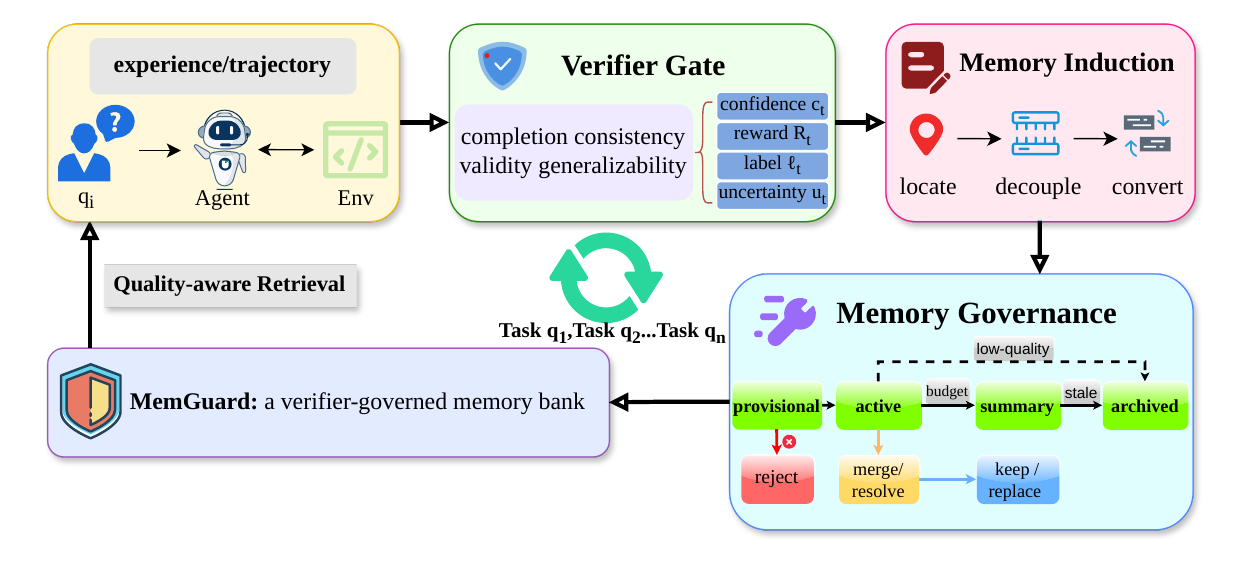}
\caption{Overview of MemGuard. MemGuard converts verifier signals into persistent descriptors attached to memory records, then uses them to govern admission, lifecycle updates, retrieval, and failure-guard injection across a task stream.}
\label{fig:memguard-overview}
\end{figure*}

\section{Related Work}
\label{sec:related-work}

\subsection{Experience Memory for LLM Agents}

Agent memory systems differ in what they store and when stored information may influence later decisions. Early systems store observations, conversational facts, or context tiers for recall \citep{park2023generativeagents,packer2023memgpt}; MemoryBank adds time-aware forgetting \citep{zhong2023memorybank}, and MemoryOS separates storage, update, retrieval, and response generation \citep{kang-etal-2025-memory}. These systems establish persistent memory, but their main unit is conversational or contextual information rather than task-level experience whose correctness must be judged after environment interaction.

Task-level experience memory converts trajectories into reusable behavior through reflections, skills, workflows, or reasoning strategies. Reflexion and ExpeL store verbal feedback or extracted insights from prior attempts \citep{shinn2023reflexion,zhao2023expel}; Voyager builds an embodied skill library from successful behavior \citep{wang2023voyager}. Synapse retrieves computer-control trajectories as exemplars \citep{zheng2024synapse}, AWM induces workflow memories from successful web trajectories \citep{wang2024agentworkflowmemory}, and ReasoningBank abstracts successful and failed trajectories into higher-level reasoning strategies \citep{ouyang2025reasoningbank}. MemGuard shifts the focus from what to remember to how a memory item should be admitted, weighted, revised, and retired.

\subsection{Verifier-Guided Selection and Supervision}

Verifier-guided methods use judgment signals to improve current outputs, reasoning paths, or trajectories. Prior work trained or prompted verifiers to select reliable answers \citep{cobbe2021trainingverifiers,xie2023selfevaluation}, used self- or tool-feedback for repair \citep{madaan2023selfrefine,chen2023selfdebugging}, trained critic models \citep{wang2023shepherd}, and analyzed LLM-as-a-Judge behavior \citep{zheng2023judging}. Complementary work studies incomplete fine-tuning and uncertainty-aware reward modeling \citep{xue2026supervised,xue2026reason}, question calibration \citep{xue2024question}, and semantic representation \citep{xue2023dual,xue2025structcoh}. In this paradigm, the verifier usually acts as a selector or critic for the current episode.

Recent work structures verifier signals through process reward models, automatic process supervision, reward-guided search, and generative reward modeling \citep{lightman2023letsverify,wang2024mathshepherd,zhang2024restmcts,zhang2024generativeverifiers}. LLM-as-a-Verifier \citep{kwok2026llmverifier} decomposes trajectory verification into multiple criteria and repeated views for one-time selection. MemGuard adopts this scoring style, derives confidence and uncertainty from score-token distributions, and stores $d_m=(R_m,c_m,\ell_m,\nu_m)$ with each record so it gates retrieval, conflict resolution, summarization, and archival. The contribution is not a new verifier, but persistent verifier-derived governance metadata.

\subsection{Memory Governance and Reliable Reuse}

Reliable long-term reuse requires more than retrieving semantically similar items. RAG retrieves external documents from a non-parametric index \citep{lewis2020retrievalaugmented}, and knowledge editing changes factual associations inside model parameters \citep{meng2022rome,meng2022memit}; MemGuard instead keeps the backbone fixed and governs an external bank of self-generated experiences. This distinction matters because an agent memory can be relevant and readable yet harmful if it came from accidental success, stale environment state, or misattributed failure.

Recent studies show that memory management affects agent reliability. MemBench evaluates memory effectiveness, efficiency, and capacity \citep{tan-etal-2025-membench}; HiAgent reduces redundant long-horizon histories \citep{hu-etal-2025-hiagent}; \citet{xiong2025memorymanagement} identify experience-following behavior; and \citet{wang-etal-2025-unveiling-privacy} show privacy risks. Complementary proposals address state-evolution attribution in long-term memory, contribution-aware retention of long-horizon logs, and semantic propagation control in multi-agent systems \citep{zhang2026memmark,liu2026conmem,dai2026safeflow}. MemGuard addresses the experience-governance side of this problem through verifier-guided admission, quality-aware retrieval, deduplication, conflict resolution, summarization, and archival. Reflexion turns prior attempts into appended verbal reflections, Synapse retrieves earlier computer-control trajectories as exemplars, and MemoryOS organizes conversational memory into storage, update, retrieval, and response tiers. These systems make memory persistent, but their write-time feedback or update signal is not a first-class lifecycle field that continues to govern later retrieval, conflict handling, summarization, and retirement. Verifier-based systems solve a complementary problem: they score or select the current answer, reasoning path, trajectory, or candidate set. LLM-as-a-Verifier \citep{kwok2026llmverifier}, for example, estimates trajectory reward from multi-criterion, repeated verification for the current decision episode. MemGuard connects these two lines by making $d_m=(R_m,c_m,\ell_m,\nu_m)$ persistent lifecycle metadata: the descriptor adjusts retrieval scores, blocks risky injection, prioritizes conflict resolution, and controls summarization or archival after the memory has been written.

\section{Method}
\label{sec:method}

\subsection{Overview}

MemGuard is a long-term experience memory system for LLM agents. Given a task stream $\mathcal{X}=\{x_1,\ldots,x_T\}$, the agent at step $t$ can access the current task, the base tool environment, and the previous memory bank $\mathcal{M}_{t-1}$, but not future tasks or human-labeled feedback. MemGuard retrieves a small set of reliable memories from $\mathcal{M}_{t-1}$, renders them as a compact memory block, and injects the block into the agent context. After the agent finishes, MemGuard records the trajectory $\tau_t$, final answer or patch $y_t$, and execution status $e_t$. A verifier scores the trajectory, and a memory induction module extracts candidate experiences. The governance module then decides whether to discard, activate, merge, replace, summarize, or archive these candidates, yielding $\mathcal{M}_t$.

Unlike systems that store complete trajectories, MemGuard distills noisy interaction histories into compact, verifiable units. It removes instance-specific details and keeps transferable decomposition patterns, environment observations, tool-use rules, and failure-avoidance principles \citep{ouyang2025reasoningbank}, so the agent learns from both successes and failures.

The key design choice is to persist verifier outputs as memory metadata rather than using them only for a one-time acceptance decision. Each memory carries a verifier descriptor $d_m=(R_m,c_m,\ell_m,\nu_m)$, where $R_m$ denotes the trajectory reward, $c_m$ denotes confidence, $\ell_m$ is the verifier label, and $\nu_m$ records verification time. This differs from a verifier-only filter: the verifier decision is not discarded after admission, but remains attached to the record and changes later retrieval, merging, conflict resolution, summarization, and archival. Admission, retrieval, conflict handling, summarization, and archival are all functions of both memory content and $d_m$.

\begin{table}[t]
\centering
\footnotesize
\setlength{\tabcolsep}{3pt}
\caption{Notation used in Section~\ref{sec:method}.}
\label{tab:notation}
\begin{tabular}{lp{0.67\columnwidth}}
\toprule
\textbf{Symbol} & \textbf{Meaning} \\
\midrule
$\mathcal{X},x_t$ & task stream and task at time $t$ \\
$\mathcal{M}_t$ & governed memory bank after task $t$ \\
$m,\mathcal{C}_t$ & memory record and induced candidate set \\
$\tau_t,y_t,e_t$ & trajectory, final output, and execution status \\
$d_m$ & verifier descriptor attached to record $m$ \\
$R_m,c_m,\ell_m,\nu_m$ & reward, confidence, verifier label, and verification time \\
$\mathcal{V},\mathcal{R}_t$ & verifier views and the view set used for trajectory $t$ \\
$\mathcal{V}_{\mathrm{score}},v,\phi(v)$ & allowed score tokens, score token, and normalized token map \\
$p_\theta(v\mid\cdot)$ & verifier score-token distribution for a criterion and view \\
$R_t,c_t,\ell_t,\nu_t$ & aggregated trajectory reward, confidence, verifier label, and verification time \\
$b_t,s_t,u_t$ & boundary score, dispersion score, and uncertainty \\
$\tilde r,\tilde q,\tilde p$ & normalized relevance, quality, and applicability signals \\
$\tilde b_{\mathrm{rec}},\tilde b_{\mathrm{use}}$ & normalized recency and successful-reuse signals \\
$\tilde h_{\mathrm{cf}},\tilde h_{\mathrm{stale}}$ & normalized conflict and staleness penalties \\
$\tilde h_{\mathrm{ver}},\tilde h_{\mathrm{ovr}}$ & normalized verifier-risk and over-generalization penalties \\
$\gamma_{\mathrm{low}},\gamma_{\mathrm{high}},\Delta_b$ & reward band and taper width for boundary trajectories \\
$\eta_{\mathrm{verify}}$ & uncertainty threshold for repeated verification \\
$\tau_{\mathrm{cf}}$ & structured-signature similarity threshold for conflict detection \\
$S^{+}(x_t,m)$ & positive retrieval score of memory $m$ for task $x_t$ \\
$S^{-}(x_t,m)$ & failure-guard retrieval score of memory $m$ for task $x_t$ \\
$\mathrm{sim}_{ij}$ & structured-signature similarity between records $i$ and $j$ \\
$B$ & active-memory budget \\
\bottomrule
\end{tabular}
\end{table}

\begin{algorithm}[t]
\caption{MemGuard update loop.}
\label{alg:memguard}
\small
\begin{algorithmic}[1]
\Require task $x_t$, memory bank $\mathcal{M}_{t-1}$
\State $\mathcal{P}_t \leftarrow \mathrm{RetrievePositive}(x_t,\mathcal{M}_{t-1})$
\State $\mathcal{G}_t \leftarrow \mathrm{RetrieveGuards}(x_t,\mathcal{M}_{t-1})$
\State $(\tau_t,y_t,e_t) \leftarrow \mathrm{Agent}(x_t,\mathcal{P}_t,\mathcal{G}_t)$
\State $\mathcal{V}\leftarrow\{\mathrm{full}\}$; compute $(R_t,u_t)$ on $\mathcal{V}$
\If{$u_t>\eta_{\mathrm{verify}}$}
  \State $\mathcal{V}\leftarrow\{\mathrm{full},\mathrm{evidence},\mathrm{risk}\}$
\EndIf
\State $(R_t,c_t,\ell_t) \leftarrow \mathrm{AggregateVerify}(\tau_t,y_t,e_t,\mathcal{V})$
\State $\mathcal{C}_t \leftarrow \mathrm{Induce}(\tau_t,R_t,c_t,\ell_t)$
\ForAll{$m\in\mathcal{C}_t$}
  \State attach $d_m=(R_m,c_m,\ell_m,\nu_m)\leftarrow(R_t,c_t,\ell_t,\nu_t)$
  \State assign initial state: active, provisional, or rejected
  \If{$m$ is failure avoidance}
    \State add $m$ to guard pool
  \EndIf
\EndFor
\State $\mathcal{M}' \leftarrow \mathrm{MergeDuplicates}(\mathcal{M}_{t-1}\cup\mathcal{C}_t)$
\State $\mathcal{M}'' \leftarrow \mathrm{ResolveConflicts}(\mathcal{M}',\text{Eq.~\ref{eq:conflict}})$
\State $\mathcal{M}_t \leftarrow \mathrm{SummarizeArchive}(\mathcal{M}'',B)$
\State \Return $\mathcal{M}_t$
\end{algorithmic}
\end{algorithm}

\subsection{Structured Memory Units}

MemGuard represents each experience as a structured record with title, content, type, lifecycle state, source status, quality, confidence, usage statistics, deduplication signature, conflict links, and verifier descriptor $d_m$. We use \textit{procedural hints}, \textit{tool-usage} memories, and \textit{failure-avoidance} memories; failure memories are rendered as constraints rather than direct action recipes. Candidate records can be \textit{provisional}, \textit{active}, \textit{summary}, or \textit{archived}; only active and selected summary records enter ordinary retrieval. The induction prompt and full schema are in Appendices~\ref{sec:induction-prompt} and~\ref{sec:memory-schema}.

\subsection{Verifier-Guided Memory Admission}

Verifier signals decide whether each candidate experience enters long-term memory. When a task finishes, the verifier receives the task, trajectory, final output, and runtime status $e_t$. The status $e_t$ contains only information the agent itself observes: exit codes, public test output, tool errors, visible web confirmations, and explicit environment messages. It never contains hidden benchmark labels, held-out evaluator outcomes, or oracle annotations. Following the multi-criterion verifier style of LLM-as-a-Verifier \citep{kwok2026llmverifier}, MemGuard scores four criteria -- task completion, evidence consistency, execution validity, and generalizability -- and aggregates the score-token distributions and rationales into a label $\ell_t \in \{\textit{verified\_success}, \textit{verified\_fail}, \textit{uncertain}\}$, trajectory reward $R_t$, and confidence $c_t$. High-uncertainty cases trigger re-verification under evidence-focused and risk-focused views of the same trajectory.

Formally, MemGuard uses $C$ verification criteria, a trajectory-specific view set $\mathcal{R}_t$, and $G$ discrete score tokens, where $G$ controls score granularity. Let $\mathcal{V}_{\mathrm{score}}=\{1,2,3,4,5\}$ by default, with $\phi(v)=(v-1)/4$ mapping each token to $[0,1]$. For task $x_t$, trajectory $\tau_t$, criterion $j$, and view $\rho\in\mathcal{R}_t$, the verifier applies a softmax over the allowed score tokens to obtain $p_\theta(v \mid x_t,j,\tau_t,\rho)$. MemGuard estimates trajectory reward as a weighted expectation over these distributions:
\begin{equation}
\label{eq:reward}
R_t =
\sum_{j=1}^{C}
\frac{w_j}{|\mathcal{R}_t|}
\sum_{\rho\in\mathcal{R}_t}
\mathrm{E}_{v\sim p_\theta(\cdot \mid x_t,j,\tau_t,\rho)}[\phi(v)],
\end{equation}
where $w_j\geq0$ and $\sum_j w_j=1$. We use uniform weights in the reported experiments, while exposing $w_j$ so domains can assign higher weight to task completion, execution validity, or generalizability when desired.
The factor $1/|\mathcal{R}_t|$ implements uniform averaging over the views actually used for trajectory $t$; ordinary trajectories use the full view only, while boundary trajectories add evidence-focused and risk-focused views. We do not use view-specific weights in the main experiments, so $w_j$ only weights criteria, not views.
The fixed ordinal support also exposes distributional uncertainty around boundary cases.

A single uncertainty score triggers repeated verification. The score combines reward proximity to the uncertainty band with disagreement among criteria or views; these signals jointly increase uncertainty rather than acting as separate hard triggers. We fix $\gamma_{\mathrm{low}}=0.45$, $\gamma_{\mathrm{high}}=0.65$, and taper width $\Delta_b=0.10$ once on a held-out development stream. Let
\[
\bar r_{t,j,\rho}
=\mathrm{E}_{v\sim p_\theta(\cdot \mid x_t,j,\tau_t,\rho)}[\phi(v)]
\]
be the mean score for criterion $j$ under view $\rho$. We compute population standard deviations for
\[
\begin{aligned}
d_t^{\mathrm{crit}}
&= \frac{1}{|\mathcal{R}_t|}
\sum_{\rho\in\mathcal{R}_t}
\operatorname{std}_{1\leq j\leq C}\bar r_{t,j,\rho},\\
d_t^{\mathrm{view}}
&= \frac{1}{C}
\sum_{j=1}^{C}
\operatorname{std}_{\rho\in\mathcal{R}_t}\bar r_{t,j,\rho}.
\end{aligned}
\]
Because $\bar r_{t,j,\rho}\in[0,1]$, the maximum population standard deviation is $\sigma_{\max}=0.5$. We normalize
\[
\begin{aligned}
\hat d_t^{\mathrm{crit}}
&=\min(1,d_t^{\mathrm{crit}}/\sigma_{\max}),\\
\hat d_t^{\mathrm{view}}
&=\min(1,d_t^{\mathrm{view}}/\sigma_{\max}).
\end{aligned}
\]
The boundary score is
\[
\begin{aligned}
a_t &=
[\gamma_{\mathrm{low}}-R_t]_+
+[R_t-\gamma_{\mathrm{high}}]_+,\\
b_t &= \max(0,1-a_t/\Delta_b),
\end{aligned}
\]
where $[z]_+=\max(z,0)$.
The dispersion score is
\[
s_t =
\begin{cases}
\hat d_t^{\mathrm{crit}}, & |\mathcal{R}_t|=1,\\
\max(\hat d_t^{\mathrm{crit}},\hat d_t^{\mathrm{view}}),
& |\mathcal{R}_t|>1.
\end{cases}
\]
This definition lets criterion-level disagreement contribute to repeated verification even before auxiliary views are computed. MemGuard then computes
\begin{equation}
\label{eq:uncertainty}
u_t = \alpha b_t + (1-\alpha)s_t,
\end{equation}
with $\alpha=0.5$ by default, giving equal weight to boundary proximity and dispersion because neither signal dominated on the held-out development stream. In our experiments, ordinary trajectories use one view and uncertain cases use up to three views. Confidence combines label agreement, reward stability, and rationale consistency; a candidate becomes active only when $R_t \geq 0.70$ and $c_t \geq 0.60$. Otherwise it is rejected, kept provisional, or routed to failure-avoidance storage. Sensitivity analyses are in Appendices~\ref{sec:score-granularity}--\ref{sec:feedback-verifier-controls}, and the aggregation protocol is in Appendix~\ref{sec:diagnostic-protocols}.

Each candidate is parsed into title, description, and content, abstracted away from instance-specific details, and tied to cited observations or failure points. Writing $R_t$, $c_t$, and $\ell_t$ into record fields prevents relevant but low-confidence experience from carrying the same influence as verified experience.

\subsection{Precision-Oriented Retrieval}

MemGuard prioritizes precision over maximum recall during retrieval. First, a hard filter removes archived records, low-relevance records, and records whose verifier label or confidence makes them unsafe for active injection. The remaining procedural and tool-usage memories are ranked by a normalized soft score:
\begin{equation}
\label{eq:positive-score}
\begin{split}
S^{+}(x_t,m) ={}&
 \beta_r \tilde r(x_t,m)
 + \beta_q \tilde q(m) \\
&+ \beta_p \tilde p(m)
 + \beta_{\mathrm{rec}}\tilde b_{\mathrm{rec}}(m) \\
&+ \beta_{\mathrm{use}}\tilde b_{\mathrm{use}}(m)
 - \beta_{\mathrm{cf}}\tilde h_{\mathrm{cf}}(m) \\
&- \beta_{\mathrm{stale}}\tilde h_{\mathrm{stale}}(m)
 - \beta_{\mathrm{ver}}\tilde h_{\mathrm{ver}}(m).
\end{split}
\end{equation}
All tilded terms are min-max normalized to $[0,1]$. The weights separate utility from risk: relevance, verifier quality, applicability, recency, and successful reuse increase the score, while conflict, staleness, and verifier risk decrease it. The resulting scores are used for filtering and top-$k$ ranking and are not clipped to $[0,1]$. They are fixed on a held-out development stream; Appendix~\ref{sec:feedback-verifier-controls} shows that uniform and random-weight variants remain close. Failure-avoidance memories use a separate guard path with stricter relevance-score and confidence thresholds and are rendered as constraints such as ``verify the stable row identifier.'' They are ranked by
\begin{equation}
\label{eq:guard-score}
\begin{split}
S^{-}(x_t,m) ={}&
 \delta_r \tilde r(x_t,m)
 + \delta_q \tilde q(m)
 + \delta_p \tilde p(m)\\
&- \delta_{\mathrm{cf}}\tilde h_{\mathrm{cf}}(m)\\
&- \delta_{\mathrm{stale}}\tilde h_{\mathrm{stale}}(m)
 - \delta_{\mathrm{ovr}}\tilde h_{\mathrm{ovr}}(m),
\end{split}
\end{equation}
where $\tilde h_{\mathrm{ovr}}$ penalizes over-general failure rules and is used only in the failure-guard path. Retrieval budgets and guard scoring details are in Appendix~\ref{sec:hyperparameters}.

\subsection{Memory Consolidation and Budgeted Governance}

MemGuard maintains the long-term memory bank through merging, replacement, summarization, and archival. For each candidate, it compares content similarity, structured signatures, and conflict links against similar active records. A signature contains the memory type, abstract task pattern, normalized action category, tool or website scope, and applicability condition. We define conflict as
\begin{equation}
\label{eq:conflict}
\begin{split}
\mathrm{conflict}(m_i,m_j)={}&
\mathbf{1}[\mathrm{sim}_{ij}>\tau_{\mathrm{cf}}] \\
&\cdot \mathbf{1}[\mathrm{act}_i\neq \mathrm{act}_j \vee \ell_i\neq \ell_j],
\end{split}
\end{equation}
where $\mathrm{sim}_{ij}$ compares structured signatures, $\tau_{\mathrm{cf}}=0.72$ is the conflict threshold, and $\mathrm{act}_i$ denotes the recommended action for procedural/tool memories or the avoided action for failure guards. Conflict checks are within type. Duplicates are discarded, complementary records are merged, and conflicts are resolved by verifier label, reward, confidence, recency, and usage. If two records match in structure and action but differ only in label, MergeDuplicates keeps the higher-confidence label when rewards are close; ResolveConflicts handles remaining reward or confidence disagreement. MemGuard also clusters repeated successes into summaries, archives stale or low-quality records, and enforces a fixed active-memory budget. Details are in Appendices~\ref{sec:governance-operations} and~\ref{sec:hyperparameters}.

\begin{table*}[!t]
\centering
\caption{Main results across terminal, software-engineering, and web-agent benchmarks. All metrics report mean $\pm$ standard deviation over five seeds. SR denotes success rate, Resolve Rate measures issue-resolving success, SSR denotes step success rate, and AS denotes average steps.}
\label{tab:terminal-swe}
\scriptsize
\setlength{\tabcolsep}{2.4pt}
\renewcommand{\arraystretch}{0.86}
\resizebox{\textwidth}{!}{%
\begin{tabular}{ccccccccc}
\toprule
\multirow{2}{*}{\textbf{Methods}}
& \multicolumn{2}{c}{\textbf{Terminal-Bench 2.0}}
& \multicolumn{2}{c}{\textbf{SWE-Bench Verified}}
& \multicolumn{2}{c}{\textbf{WebArena}}
& \multicolumn{2}{c}{\textbf{Mind2Web}} \\
\cmidrule(lr){2-3}\cmidrule(lr){4-5}\cmidrule(lr){6-7}\cmidrule(lr){8-9}
& SR $\uparrow$ & AS $\downarrow$
& Resolve Rate $\uparrow$ & AS $\downarrow$
& SR $\uparrow$ & AS $\downarrow$
& SSR $\uparrow$ & AS $\downarrow$ \\
\midrule
\rowcolor{modelrow}\multicolumn{9}{c}{\small\textbf{\textit{Qwen-3.5-Flash}}} \\
No Memory & $48.5{\pm}1.6$ & $42.8{\pm}2.7$ & $67.0{\pm}1.1$ & $45.6{\pm}1.4$ & $32.8{\pm}1.0$ & $11.9{\pm}0.5$ & $33.2{\pm}0.7$ & $18.9{\pm}0.6$ \\
Synapse & $49.2{\pm}0.9$ & $42.6{\pm}1.8$ & $66.7{\pm}1.4$ & $45.8{\pm}0.9$ & $33.5{\pm}1.2$ & $11.6{\pm}0.4$ & $33.9{\pm}1.0$ & $18.6{\pm}0.5$ \\
AWM & $50.1{\pm}1.3$ & $41.8{\pm}2.2$ & $68.4{\pm}0.8$ & $44.2{\pm}1.6$ & $35.3{\pm}0.9$ & $11.2{\pm}0.3$ & $34.4{\pm}0.5$ & $18.3{\pm}0.6$ \\
ReasoningBank & $52.8{\pm}2.0$ & $40.9{\pm}1.5$ & $70.8{\pm}1.5$ & $42.5{\pm}1.1$ & $38.6{\pm}1.4$ & $10.4{\pm}0.6$ & $37.2{\pm}0.9$ & $17.2{\pm}0.4$ \\
Verifier-only Filter & $53.0{\pm}1.1$ & $40.7{\pm}3.0$ & $71.9{\pm}1.0$ & $41.5{\pm}1.7$ & $42.6{\pm}0.8$ & $\phantom{0}9.1{\pm}0.5$ & $37.8{\pm}0.6$ & $15.4{\pm}0.5$ \\
\rowcolor{bestrow}\textbf{MemGuard} & \textbf{\boldmath$55.8{\pm}1.8$} & \textbf{\boldmath$38.8{\pm}2.5$} & \textbf{\boldmath$73.4{\pm}1.2$} & \textbf{\boldmath$39.9{\pm}1.0$} & \textbf{\boldmath$44.2{\pm}1.1$} & \textbf{\boldmath$\phantom{0}8.7{\pm}0.3$} & \textbf{\boldmath$40.7{\pm}1.1$} & \textbf{\boldmath$13.8{\pm}0.6$} \\
\midrule
\rowcolor{modelrow}\multicolumn{9}{c}{\small\textbf{\textit{Qwen-3.5-Plus}}} \\
No Memory & $60.8{\pm}1.0$ & $34.2{\pm}1.9$ & $76.6{\pm}1.3$ & $35.0{\pm}0.8$ & $44.2{\pm}0.9$ & $\phantom{0}9.8{\pm}0.4$ & $42.3{\pm}0.8$ & $16.4{\pm}0.5$ \\
Synapse & $61.4{\pm}1.7$ & $34.9{\pm}2.4$ & $77.4{\pm}0.7$ & $35.6{\pm}1.6$ & $45.0{\pm}1.3$ & $\phantom{0}9.9{\pm}0.5$ & $42.9{\pm}0.4$ & $16.1{\pm}0.6$ \\
AWM & $60.6{\pm}0.8$ & $34.6{\pm}3.1$ & $78.2{\pm}1.5$ & $34.3{\pm}1.1$ & $46.5{\pm}1.0$ & $\phantom{0}9.2{\pm}0.6$ & $43.4{\pm}0.9$ & $15.9{\pm}0.4$ \\
ReasoningBank & $64.6{\pm}2.1$ & $32.1{\pm}1.4$ & $80.1{\pm}0.9$ & $32.7{\pm}1.5$ & $50.5{\pm}1.2$ & $\phantom{0}8.5{\pm}0.4$ & $46.2{\pm}0.7$ & $14.8{\pm}0.6$ \\
Verifier-only Filter & $64.8{\pm}1.4$ & $32.4{\pm}2.8$ & $81.9{\pm}1.1$ & $31.6{\pm}0.9$ & $52.0{\pm}1.5$ & $\phantom{0}8.4{\pm}0.5$ & $50.0{\pm}1.0$ & $11.9{\pm}0.5$ \\
\rowcolor{bestrow}\textbf{MemGuard} & \textbf{\boldmath$67.4{\pm}1.9$} & \textbf{\boldmath$30.4{\pm}1.7$} & \textbf{\boldmath$83.6{\pm}1.6$} & \textbf{\boldmath$30.6{\pm}1.2$} & \textbf{\boldmath$58.4{\pm}0.8$} & \textbf{\boldmath$\phantom{0}6.9{\pm}0.3$} & \textbf{\boldmath$51.8{\pm}0.5$} & \textbf{\boldmath$11.5{\pm}0.4$} \\
\midrule
\rowcolor{modelrow}\multicolumn{9}{c}{\small\textbf{\textit{Gemini-3-Flash}}} \\
No Memory & $55.6{\pm}1.5$ & $37.4{\pm}2.6$ & $74.2{\pm}0.8$ & $40.7{\pm}1.4$ & $39.4{\pm}1.1$ & $10.7{\pm}0.6$ & $36.7{\pm}0.6$ & $17.2{\pm}0.5$ \\
Synapse & $55.3{\pm}0.9$ & $37.8{\pm}1.3$ & $75.1{\pm}1.2$ & $40.0{\pm}1.7$ & $40.2{\pm}0.8$ & $10.4{\pm}0.4$ & $37.2{\pm}1.1$ & $16.9{\pm}0.6$ \\
AWM & $56.8{\pm}1.6$ & $36.4{\pm}2.9$ & $74.6{\pm}1.0$ & $40.6{\pm}0.9$ & $42.0{\pm}1.4$ & $10.0{\pm}0.3$ & $37.8{\pm}0.8$ & $16.6{\pm}0.5$ \\
ReasoningBank & $59.0{\pm}1.2$ & $35.6{\pm}1.1$ & $78.0{\pm}1.5$ & $37.9{\pm}1.3$ & $45.4{\pm}0.9$ & $\phantom{0}9.3{\pm}0.5$ & $39.5{\pm}0.4$ & $15.5{\pm}0.4$ \\
Verifier-only Filter & $60.3{\pm}2.0$ & $34.9{\pm}1.8$ & $77.7{\pm}0.7$ & $38.4{\pm}1.6$ & $49.6{\pm}1.3$ & $\phantom{0}8.2{\pm}0.4$ & $40.1{\pm}1.0$ & $13.5{\pm}0.6$ \\
\rowcolor{bestrow}\textbf{MemGuard} & \textbf{\boldmath$61.8{\pm}1.7$} & \textbf{\boldmath$34.1{\pm}2.3$} & \textbf{\boldmath$80.4{\pm}1.4$} & \textbf{\boldmath$36.1{\pm}1.0$} & \textbf{\boldmath$50.7{\pm}1.5$} & \textbf{\boldmath$\phantom{0}8.1{\pm}0.3$} & \textbf{\boldmath$43.8{\pm}0.7$} & \textbf{\boldmath$12.4{\pm}0.5$} \\
\midrule
\rowcolor{modelrow}\multicolumn{9}{c}{\small\textbf{\textit{Gemini-3.1-Pro}}} \\
No Memory & $68.4{\pm}0.8$ & $30.7{\pm}2.7$ & $78.2{\pm}1.3$ & $28.5{\pm}1.2$ & $56.4{\pm}1.0$ & $\phantom{0}8.1{\pm}0.4$ & $46.4{\pm}0.9$ & $14.9{\pm}0.6$ \\
Synapse & $69.2{\pm}1.4$ & $31.2{\pm}1.6$ & $77.8{\pm}0.7$ & $29.3{\pm}1.5$ & $57.5{\pm}0.8$ & $\phantom{0}8.0{\pm}0.5$ & $47.1{\pm}0.6$ & $14.6{\pm}0.4$ \\
AWM & $70.1{\pm}2.1$ & $30.1{\pm}2.0$ & $79.9{\pm}1.1$ & $28.0{\pm}0.8$ & $59.5{\pm}1.4$ & $\phantom{0}7.5{\pm}0.3$ & $47.6{\pm}1.1$ & $14.3{\pm}0.5$ \\
ReasoningBank & $73.0{\pm}1.0$ & $28.8{\pm}1.3$ & $82.2{\pm}1.6$ & $26.8{\pm}1.7$ & $63.1{\pm}0.9$ & $\phantom{0}7.0{\pm}0.4$ & $50.2{\pm}0.7$ & $13.2{\pm}0.6$ \\
Verifier-only Filter & $73.2{\pm}1.8$ & $28.9{\pm}3.1$ & $83.3{\pm}0.9$ & $26.2{\pm}1.0$ & $67.8{\pm}1.2$ & $\phantom{0}6.1{\pm}0.6$ & $51.0{\pm}0.5$ & $11.0{\pm}0.4$ \\
\rowcolor{bestrow}\textbf{MemGuard} & \textbf{\boldmath$75.7{\pm}1.3$} & \textbf{\boldmath$27.3{\pm}2.5$} & \textbf{\boldmath$85.3{\pm}1.2$} & \textbf{\boldmath$25.4{\pm}1.4$} & \textbf{\boldmath$69.1{\pm}1.5$} & \textbf{\boldmath$\phantom{0}5.9{\pm}0.3$} & \textbf{\boldmath$55.5{\pm}1.0$} & \textbf{\boldmath$10.4{\pm}0.5$} \\
\bottomrule
\end{tabular}
}
\renewcommand{\arraystretch}{1.0}
\end{table*}

\section{Experiments}

\subsection{Experimental Setup}

We evaluate on Terminal-Bench 2.0 \citep{terminalbench2026}, SWE-Bench Verified \citep{jimenez2024swebench}, WebArena \citep{zhou2024webarena}, and Mind2Web \citep{deng2023mind2web} using Qwen-3.5-Flash, Qwen-3.5-Plus \citep{qwen2026qwen35}, Gemini-3-Flash, and Gemini-3.1-Pro \citep{googledeepmind2025gemini3flash,googledeepmind2026gemini31pro}. We compare No Memory, Synapse \citep{zheng2024synapse}, AWM \citep{wang2024agentworkflowmemory}, ReasoningBank \citep{ouyang2025reasoningbank}, Verifier-only Filter, and MemGuard.

Each benchmark-model run is a continuous task stream: the memory bank is reset once, then updated after every task. All methods share the same runtime, task order, step budget, decoding, retriever, injected-memory limit, and memory-context budget.

No method uses hidden benchmark gold labels or final evaluator outcomes during memory admission; memory updates use only the logged trajectory, final answer or patch, visible tool outputs, and runtime feedback available to the agent process. AS denotes the average number of agent action steps per task, not API calls. Appendix~\ref{sec:experiment-details} gives benchmark, baseline, model-version, and prompt details.

\subsection{Main Results}

Table~\ref{tab:terminal-swe} reports the main results. MemGuard achieves the highest primary success metric and the lowest AS in every backbone--benchmark setting. Gains are largest on interactive web tasks, where unstable navigation states and task-specific observations often pollute memory. On terminal and software-engineering tasks, MemGuard still gains by filtering low-quality experience and cutting repeated exploration.

Across all 16 backbone--benchmark settings, MemGuard achieves the best success metric. WebArena and Mind2Web show the largest gains: web agents face unstable page state, ambiguous element identity, and misleading partial successes; Terminal-Bench 2.0 and SWE-Bench Verified show 2.4--3.5 point gains. Paired bootstrap tests on task-level outcomes in Appendix~\ref{sec:significance-tests} remain significant for all 16 settings after Benjamini--Hochberg false-discovery-rate correction ($q=0.004$--$0.032$); seven remain significant under the stricter Holm family-wise correction. The aggregate ordering is not universal at the seed level: across 320 matched success-metric and AS comparisons against ReasoningBank and Verifier-only Filter, MemGuard has 11 losses and one tie. The strict Mind2Web task-level SR is intentionally conservative, so we use SSR as the primary Mind2Web summary metric.

The standard deviations in Table~\ref{tab:terminal-swe} summarize five seeded runs with the same benchmark split, memory budget, retrieval budget, and decoding configuration; the main ranking is stable under run-level variation.

Appendix~\ref{sec:detailed-results} gives split-level WebArena and Mind2Web results, and Figure~\ref{fig:accumulation} shows that MemGuard's advantage emerges later in the same continuous task streams rather than from isolated per-task resets. A separate chronological stream of all 2,294 SWE-bench Full issues extends the Verified horizon by $4.6\times$: MemGuard improves Resolve Rate by 2.96 points and reduces AS by 2.10 steps over ReasoningBank, with the gap widening in later stream windows (Appendix~\ref{sec:swe-full-long-horizon}).

\subsection{Verifier Control and Overhead}
\label{sec:verifier-control}

The verifier-only control improves the primary success metric over ReasoningBank in 15 of 16 backbone--benchmark settings, showing that trajectory verification helps, but MemGuard achieves higher success and lower AS than this control in all 16 settings. Appendix Table~\ref{tab:overhead} reports prompt tokens, verifier/governance tokens, latency, and storage: MemGuard adds 1.0--2.5k tokens and 5.1--15.0s per task over the no-memory reference, but stays below the other memory and verifier baselines by reducing agent steps.

Appendix Table~\ref{tab:verifier-calibration} audits verifier quality, with 86\% agreement against author inspection for Qwen-3.5-Plus and 83\% for Qwen-3.5-Flash. An independent post-hoc check joins 51,896 of 52,280 frozen task-completion decisions to benchmark-provided outcomes that were unavailable to every online component, obtaining 88.1\% macro balanced accuracy and a 9.3\% false-positive rate. Table~\ref{tab:false-accept-tracing} shows that governance blocks 38 of 60 false accepts before active-memory promotion. Appendix Table~\ref{tab:verifier-noise} further stress-tests verifier dependence: at 30\% simulated label noise, MemGuard loses 0.3--2.6 fewer metric points than the verifier-only control. This gap is consistent with the mechanism in Section~\ref{sec:verifier-control-details}: provisional states, confidence-aware retrieval, conflict checks, and failure-guard separation reduce corrupted labels entering active positive context. Cross-family replacement in both directions also preserves MemGuard's advantage over matched ReasoningBank in all eight tested agent--benchmark cells (Appendix Table~\ref{tab:cross-family-verifier}).

Additional controls in Appendices~\ref{sec:score-granularity}--\ref{sec:feedback-verifier-controls} show that the default score-token scale, admission thresholds, verification views, runtime-feedback setting, retrieval weights, and same-family verifier choice are not narrow sources of the result.

\subsection{Ablation Study}

We ablate verifier admission (\textit{w/o Adm.}), repeated verification (\textit{w/o Rep.}), governance (\textit{w/o Gov.}), failure memories (\textit{w/o Fail.}), and semantic-only retrieval (\textit{Sem. Ret.}). Table~\ref{tab:main-ablation} gives the compact view; Appendix~\ref{sec:ablation-details} reports all backbones. Dropping governance causes the largest success drop because duplicate, stale, and conflicting records stay active; dropping admission is the second-largest loss, and dropping failure guards mainly inflates AS. Semantic-only retrieval performs worse in every setting, showing that retrieval without verifier-quality and conflict signals cannot recover the gains. A simplified $\{\text{label},\text{confidence}\}$ descriptor retains most of the benefit but trails the full descriptor by 0.7--1.5 success-metric points across four Qwen-3.5-Plus settings (Appendix~\ref{sec:descriptor-simplification}).

Appendix~\ref{sec:robustness-governance} reports memory-health statistics, budget sensitivity, failure-memory risk analysis, a case study, and the paired bootstrap tests summarized above. In a 200-case audit, most risky failure-derived memories are rejected, kept provisional, merged into safer summaries, or archived rather than injected as action plans.

\begin{table}[!htbp]
\centering
\footnotesize
\setlength{\tabcolsep}{2.1pt}
\caption{Compact ablation on Qwen-3.5-Plus across four benchmarks. Each cell reports success metric / AS: SR for Terminal-Bench 2.0 and WebArena, Resolve Rate for SWE-Bench Verified, and SSR for Mind2Web. Sem. Ret. denotes semantic-only retrieval. Full results across four backbones are in Appendix~\ref{sec:ablation-details}.}
\label{tab:main-ablation}
\resizebox{\columnwidth}{!}{%
\begin{tabular}{lcccc}
\toprule
\textbf{Variant}
& \textbf{Terminal-Bench 2.0}
& \textbf{SWE-Bench}
& \textbf{WebArena}
& \textbf{Mind2Web} \\
\midrule
\rowcolor{bestrow}\textbf{Full} & \textbf{67.4/30.4} & \textbf{83.6/30.6} & \textbf{58.4/6.9} & \textbf{51.8/11.5} \\
w/o Adm. & 64.4/32.0 & 80.9/32.5 & 52.8/8.0 & 48.0/13.1 \\
w/o Rep. & 65.3/31.5 & 81.5/31.7 & 55.6/7.4 & 50.0/12.2 \\
w/o Gov. & 63.7/32.6 & 79.8/33.0 & 52.0/8.5 & 46.8/13.8 \\
w/o Fail. & 66.1/31.4 & 82.2/31.5 & 56.5/7.8 & 50.7/12.8 \\
Sem. Ret. & 63.4/32.9 & 79.4/33.2 & 50.1/8.8 & 45.8/14.1 \\
\bottomrule
\end{tabular}
}
\end{table}
\FloatBarrier
\section{Conclusion}

We introduced MemGuard, a verifier-guided memory governance framework for long-running LLM agents. The central idea is that experience memory should not only be retrieved but also maintained after it is written. MemGuard turns trajectory verification into persistent memory metadata: reward, confidence, verifier labels, usage statistics, and conflict links guide admission, retrieval, merging, summarization, and archival throughout a task stream.

Across Terminal-Bench 2.0, SWE-Bench Verified, WebArena, and Mind2Web, MemGuard outperforms ReasoningBank, the strongest prior baseline among the memory methods we evaluate, in all 16 backbone--benchmark settings. Its lower AS also translates into lower deployment cost relative to ReasoningBank, with 3.0--8.4k fewer total tokens and 6.5--19.0s lower latency per task. Gains are largest on web-agent benchmarks (5.3--7.9 SR points on WebArena and 3.5--5.6 SSR points on Mind2Web over ReasoningBank); on terminal and software-engineering benchmarks, MemGuard gains 2.4--3.5 points. The verifier-only control confirms that verification itself helps, while the gap to MemGuard demonstrates that one-time filtering is not the main driver: persistent verifier signals as memory attributes down-weight risky experience, retain useful failure-avoidance guards, and retire stale or conflicting records.

The broader takeaway is that reliable agent memory is a lifecycle problem. In interactive environments, harmful memories often look semantically relevant, so stronger retrieval alone is insufficient. Future work should test verifier transfer across more model families, extend memory-drift evaluation beyond the 2,294-issue stream, and study multimodal and privacy-sensitive agent settings.

\section*{Limitations}

\paragraph{Verifier dependence.}
In a manual audit of 800 verifier decisions sampled from the four benchmarks under Qwen-3.5-Plus and Qwen-3.5-Flash, verifier labels agreed with author inspection in 86\% and 83\% of cases, respectively, with disagreements driven mainly by incomplete logs or ambiguous task-success criteria. The independent benchmark-outcome check covers task completion at much larger scale, but does not replace human assessment of evidence consistency, execution validity, or generalizability. False accepts can still promote unsupported experience into active memory, and the resulting errors may accumulate in long task streams. Table~\ref{tab:false-accept-tracing} traces these false accepts through later retrieval and action influence. The noisy-verifier analysis in Table~\ref{tab:verifier-noise} tests the same failure mode by corrupting verifier labels; it suggests that lifecycle governance can buffer moderate label noise relative to verifier-only filtering, but it is not a guarantee under calibrated, systematic, or adversarial verifier errors. Stricter activation thresholds, provisional states, repeated verification, and failure-guard separation reduce this risk but do not guarantee that all harmful experiences are filtered. Cross-family verification preserves the reported advantage in the eight tested cells, but larger sweeps over verifier families, calibration sets, and human-audited domains are still needed.

\paragraph{Generalizability beyond reported streams.}
The main table reports mean success metrics and standard deviations over five seeds for each model--benchmark setting, supplemented by bootstrap tests over benchmark tasks. Larger multi-seed sweeps would give tighter effect-size estimates, especially on Terminal-Bench 2.0 and SWE-Bench Verified where task heterogeneity is high. MemGuard also uses fixed thresholds, weights, and memory budgets selected on a small held-out development stream. The sensitivity analyses show that nearby settings preserve performance on the tested web benchmarks, but new domains may require recalibration. The 2,294-issue SWE-bench Full stream is substantially longer than the main benchmarks but remains shorter and less heterogeneous than open-ended production deployments; budget saturation, lossy summaries, and stale failure guards therefore remain open issues.

\paragraph{Scope of evaluation.}
MemGuard governs external memory rather than updating backbone model parameters, so it cannot compensate when the base agent lacks a required tool-use or reasoning capability. We evaluate terminal, software-engineering, and web-agent environments, but not multimodal agents, open-world embodied agents, or production settings where incorrect memory can have legal, financial, medical, or security consequences. We compare against trajectory-, workflow-, and reasoning-memory baselines, but do not run every reflection or skill-library system because several are designed for different interaction loops. MemGuard also focuses on correctness-oriented governance; it does not yet provide a complete privacy layer for detecting, redacting, or auditing sensitive information stored in memory.

\section*{Ethical Considerations}

MemGuard is intended to improve long-running LLM-agent reliability, but memory governance introduces ethical risks that should be considered.

\paragraph{Privacy of stored trajectories.}
Agent trajectories can contain sensitive information from web pages, repositories, terminal outputs, or user instructions. Our experiments use benchmark tasks rather than private user data, but the current system does not provide a complete privacy-preserving memory layer; deployments should add redaction, access control, audit logs, and deletion mechanisms before deploying persistent memory for real users.

\paragraph{Residual verifier risk.}
Verifier-guided admission can reduce harmful experience reuse but cannot eliminate it. A verifier may falsely accept an unsupported trajectory, falsely reject a useful one, or encode biases from its model family and prompt. MemGuard mitigates this risk through provisional states, stricter activation thresholds, repeated verification, failure-guard separation, conflict resolution, and archival, but high-risk settings should require human review and rollback mechanisms.

\paragraph{Scope of safety claims.}
The reported gains in controlled benchmarks should not be read as evidence that autonomous agents are safe for unconstrained deployment. MemGuard is a reliability layer for research agents and benchmarked task streams, not a substitute for domain-specific safety evaluation, user consent, privacy review, or operational monitoring.

\section*{Acknowledgments}

We thank the anonymous reviewers, action editor, and program committee for their constructive feedback, which helped improve the empirical validation and presentation of this work. We also thank our institutions and colleagues for providing supportive research environments and helpful discussions. Generative AI tools were used for language polishing, organization, and LaTeX formatting; the authors reviewed the resulting text and take full responsibility for the methods, experiments, claims, and presentation.

\bibliography{verified_refs}

\appendix

\clearpage
\section*{Contents of Appendix}
\noindent\rule{\columnwidth}{0.4pt}
\vspace{0.4em}
\begingroup
\small
\newcommand{\appentry}[3]{%
  \par\noindent\textbf{\hyperref[#2]{#1\quad #3}}\dotfill\textbf{\pageref{#2}}\par\vspace{0.15em}}
\newcommand{\appsubentry}[3]{%
  \par\noindent\hspace{1.2em}\hyperref[#2]{#1\quad #3}\dotfill\pageref{#2}\par}
\appentry{A}{sec:experiment-details}{Experiment Details}
\appsubentry{A.1}{sec:benchmarks-splits}{Benchmarks and Splits}
\appsubentry{A.2}{sec:agent-runtime}{Agent Runtime and Decoding}
\appsubentry{A.3}{sec:baseline-implementations}{Baseline Implementations}
\appsubentry{A.4}{sec:memory-verifier-settings}{Memory and Verifier Settings}
\appsubentry{A.5}{sec:detailed-results}{Detailed Web Results}
\vspace{0.25em}
\appentry{B}{sec:prompt-templates}{Prompt Templates}
\appsubentry{B.1}{sec:verifier-prompt}{Verifier Prompt}
\appsubentry{B.2}{sec:verification-views}{Verification Views}
\appsubentry{B.3}{sec:induction-prompt}{Memory Induction Prompt}
\appsubentry{B.4}{sec:memory-injection-prompt}{Memory Injection Prompt}
\appsubentry{B.5}{sec:prompt-example}{Prompt Example}
\vspace{0.25em}
\appentry{C}{sec:efficiency-ablation}{Additional Experimental Analyses}
\appsubentry{C.1}{sec:accumulation}{Long-Horizon Task-Stream Accumulation}
\appsubentry{C.2}{sec:swe-full-long-horizon}{Extended SWE-bench Full Stream}
\appsubentry{C.3}{sec:verifier-control-details}{Verifier-Only Control and Deployment Cost}
\appsubentry{C.4}{sec:verifier-reliability}{Verifier Reliability and Label Noise}
\appsubentry{C.5}{sec:score-granularity}{Score-Token Granularity}
\appsubentry{C.6}{sec:descriptor-simplification}{Descriptor Simplification}
\appsubentry{C.7}{sec:threshold-sensitivity}{Admission Threshold and View Sensitivity}
\appsubentry{C.8}{sec:feedback-verifier-controls}{Runtime Feedback, Weight, and Cross-Family Verifier Results}
\appsubentry{C.9}{sec:ablation-details}{Ablation Details}
\vspace{0.25em}
\appentry{D}{sec:implementation-details}{Implementation and Audit Protocols}
\appsubentry{D.1}{sec:memory-schema}{Memory Record Schema}
\appsubentry{D.2}{sec:verifier-criteria}{Verifier Criteria}
\appsubentry{D.3}{sec:governance-operations}{Governance Operations}
\appsubentry{D.4}{sec:hyperparameters}{Hyperparameters and Implementation Details}
\appsubentry{D.5}{sec:diagnostic-protocols}{Diagnostic Analysis Protocols}
\appsubentry{D.6}{sec:memory-induction-example}{Memory Induction Example}
\vspace{0.25em}
\appentry{E}{sec:robustness-governance}{Robustness and Governance Analyses}
\appsubentry{E.1}{sec:governance-analysis}{Memory Health and Governance Analysis}
\appsubentry{E.2}{sec:budget-sensitivity}{Memory Budget Sensitivity}
\appsubentry{E.3}{sec:failure-risk}{Failure-Memory Risk Analysis}
\appsubentry{E.4}{sec:case-study}{Case Study}
\appsubentry{E.5}{sec:significance-tests}{Significance Tests}
\vspace{0.25em}
\appentry{F}{sec:reproducibility}{Reproducibility Statement}
\appentry{G}{sec:llm-use}{Use of LLMs}
\endgroup
\clearpage

\section{Experiment Details}
\label{sec:appendix}
\label{sec:experiment-details}

This appendix fixes the shared evaluation setup before reporting split-level web results. It first specifies benchmark scopes and runtime settings, then clarifies the baseline implementations and memory/verifier configuration used throughout the paper.

\subsection{Benchmarks and Splits}
\label{sec:benchmarks-splits}

Table~\ref{tab:benchmark-settings} summarizes the evaluation suites used in this paper. We keep each benchmark's official task definition and evaluation protocol, and reset the memory bank at the beginning of each benchmark-model run. For WebArena and Mind2Web, we additionally report split-level results in Appendix~\ref{sec:detailed-results} because their domain structure is central to understanding where governed memory helps. Counts in parentheses are official benchmark instances/tasks; WebArena overall SR uses the task-count-weighted micro-average over its five sites.

\begin{table*}[!b]
\centering
\footnotesize
\setlength{\tabcolsep}{3pt}
\caption{Benchmark settings. AS denotes the average number of agent--environment interaction steps per task.}
\label{tab:benchmark-settings}
\begin{tabular*}{\textwidth}{@{\extracolsep{\fill}}p{0.18\textwidth}p{0.20\textwidth}p{0.43\textwidth}p{0.13\textwidth}}
\toprule
\textbf{Benchmark} & \textbf{Task scope} & \textbf{Splits or domains} & \textbf{Metrics} \\
\midrule
Terminal-Bench 2.0 & Command-line operation & Official task suite & SR, AS \\
SWE-Bench Verified & Repository issue resolution & Verified issue set & Resolve Rate, AS \\
WebArena & Interactive web tasks & Shopping (187), Admin (182), GitLab (180), Reddit (106), Multi (29); Overall (684) & SR, AS \\
Mind2Web & Web-agent generalization & Cross-Task (252), Cross-Website (177), Cross-Domain (912) & EA, AF1, SSR, SR, AS \\
\bottomrule
\end{tabular*}
\vspace{0.75em}
\small
\setlength{\tabcolsep}{4pt}
\caption{Implementation comparison among baselines.}
\label{tab:baseline-implementation}
\begin{tabular*}{\textwidth}{@{\extracolsep{\fill}}llll}
\toprule
\textbf{Method} & \textbf{Stored experience} & \textbf{Verifier use} & \textbf{Lifecycle governance} \\
\midrule
No Memory & None & None & None \\
Synapse & Retrieved trajectories & None & Append-only cache \\
AWM & Workflow memories & None & Workflow-level update \\
ReasoningBank & Reasoning memories & Success/failure signal & Minimal consolidation \\
Verifier-only Filter & Reasoning memories & Admission filtering & No verifier-aware retrieval \\
MemGuard & Governed memory records & Admission, retrieval, and updates & Merge, summarize, resolve, archive \\
\bottomrule
\end{tabular*}
\end{table*}

\subsection{Agent Runtime and Decoding}
\label{sec:agent-runtime}

All methods use the same ReAct-style agent interface, tool environment, observation format, and stop criteria within each benchmark. A step is counted when the agent observes the environment, produces a thought or plan, and executes an action. WebArena uses a maximum of 30 interaction steps per task, Mind2Web follows the benchmark-provided step horizon, and Terminal-Bench 2.0 and SWE-Bench Verified use a maximum of 60 action steps. Agent decoding uses temperature 0.2 and top-$p=0.95$ for all compared methods. The verifier and governance modules are outside the environment loop, so verifier calls are not counted in AS; their cost is reported separately in Appendix~\ref{sec:verifier-control-details}.

\subsection{Baseline Implementations}
\label{sec:baseline-implementations}

Table~\ref{tab:baseline-implementation} specifies what is shared and what differs across methods. This separation is important because MemGuard is not meant to benefit from a stronger retriever, larger context budget, or more favorable task order. Synapse, ReasoningBank, Verifier-only Filter, and MemGuard use the same retrieval pool size, injected-memory budget, and memory block template; they differ in what is stored and how memory quality affects admission, retrieval, and governance.

\subsection{Memory and Verifier Settings}
\label{sec:memory-verifier-settings}

The memory bank is updated after every completed task. Candidate memories are induced from the full trajectory, final answer or patch, execution status, and verifier output. Positive memories are eligible for top-$k$ retrieval only after passing activation thresholds; failure memories are routed through the negative-guard path and appended as constraints when they pass a stricter relevance and confidence filter. Unless otherwise specified, all hyperparameters are fixed across models and benchmarks, as listed in Appendix~\ref{sec:hyperparameters}.

\subsection{Detailed Web Results}
\label{sec:detailed-results}

This section expands the web results from Table~\ref{tab:terminal-swe}. WebArena is reported by site because navigation errors and memory failures differ across shopping, administration, code-hosting, forum, and multi-site tasks. Mind2Web is reported by official split because cross-task, cross-website, and cross-domain settings stress different forms of memory transfer.

\begin{table*}[!t]
\centering
\small
\setlength{\tabcolsep}{3.6pt}
\caption{Detailed results on WebArena. Results are averaged over five seeds. SR denotes task success rate; AS is defined in Section~\ref{sec:agent-runtime}. Overall SR is the task-count-weighted micro-average over the five sites; Overall AS is the site-level average used in Table~\ref{tab:terminal-swe}.}
\label{tab:webarena-detail}
\begin{tabular}{lcccccccccccc}
\toprule
& \multicolumn{2}{c}{\shortstack{\textbf{Shopping}\\(187)}} & \multicolumn{2}{c}{\shortstack{\textbf{Admin}\\(182)}} & \multicolumn{2}{c}{\shortstack{\textbf{GitLab}\\(180)}} & \multicolumn{2}{c}{\shortstack{\textbf{Reddit}\\(106)}} & \multicolumn{2}{c}{\shortstack{\textbf{Multi}\\(29)}} & \multicolumn{2}{c}{\shortstack{\textbf{Overall}\\(684)}} \\
\cmidrule(lr){2-3}\cmidrule(lr){4-5}\cmidrule(lr){6-7}\cmidrule(lr){8-9}\cmidrule(lr){10-11}\cmidrule(lr){12-13}
\textbf{Methods} & SR & AS & SR & AS & SR & AS & SR & AS & SR & AS & SR & AS \\
\midrule
\rowcolor{modelrow}\multicolumn{13}{c}{\textbf{\textit{Qwen-3.5-Flash}}} \\
No Memory & 31.0 & 11.4 & 36.5 & 12.1 & 27.8 & 14.6 & 45.3 & 8.2 & 6.9 & 11.8 & 32.8 & 11.9 \\
Synapse & 31.8 & 11.0 & 37.1 & 12.4 & 28.6 & 14.1 & 45.8 & 8.5 & 6.9 & 11.5 & 33.5 & 11.6 \\
AWM & 33.7 & 10.8 & 39.1 & 11.4 & 30.4 & 13.8 & 48.7 & 7.9 & 3.4 & 10.5 & 35.3 & 11.2 \\
ReasoningBank & 36.9 & 10.1 & 42.0 & 10.7 & 33.1 & 13.0 & 52.8 & 7.3 & 10.3 & 10.1 & 38.6 & 10.4 \\
\rowcolor{bestrow}\textbf{MemGuard} & \textbf{42.2} & \textbf{8.7} & \textbf{48.4} & \textbf{9.1} & \textbf{38.0} & \textbf{11.2} & \textbf{58.5} & \textbf{6.4} & \textbf{17.2} & \textbf{8.2} & \textbf{44.2} & \textbf{8.7} \\
\midrule
\rowcolor{modelrow}\multicolumn{13}{c}{\textbf{\textit{Qwen-3.5-Plus}}} \\
No Memory & 41.2 & 9.4 & 48.0 & 10.1 & 38.9 & 11.7 & 60.4 & 6.8 & 13.8 & 9.2 & 44.2 & 9.8 \\
Synapse & 42.0 & 9.8 & 48.8 & 9.7 & 39.6 & 11.9 & 61.3 & 6.5 & 13.8 & 9.5 & 45.0 & 9.9 \\
AWM & 43.7 & 8.9 & 50.2 & 9.5 & 41.2 & 10.9 & 64.2 & 6.4 & 10.3 & 8.3 & 46.5 & 9.2 \\
ReasoningBank & 47.1 & 8.2 & 54.9 & 8.8 & 44.0 & 10.1 & 68.9 & 5.9 & 17.2 & 7.9 & 50.5 & 8.5 \\
\rowcolor{bestrow}\textbf{MemGuard} & \textbf{55.1} & \textbf{6.5} & \textbf{63.2} & \textbf{7.0} & \textbf{51.2} & \textbf{8.3} & \textbf{76.4} & \textbf{4.8} & \textbf{27.6} & \textbf{6.2} & \textbf{58.4} & \textbf{6.9} \\
\midrule
\rowcolor{modelrow}\multicolumn{13}{c}{\textbf{\textit{Gemini-3-Flash}}} \\
No Memory & 37.4 & 10.2 & 43.1 & 10.9 & 34.0 & 12.8 & 53.8 & 7.4 & 10.3 & 10.0 & 39.4 & 10.7 \\
Synapse & 38.2 & 9.9 & 43.9 & 11.3 & 34.8 & 12.4 & 54.5 & 7.8 & 10.3 & 9.8 & 40.2 & 10.4 \\
AWM & 40.0 & 9.7 & 45.5 & 10.1 & 36.9 & 11.8 & 57.5 & 7.0 & 6.9 & 9.1 & 42.0 & 10.0 \\
ReasoningBank & 43.1 & 9.0 & 49.0 & 9.4 & 39.7 & 11.1 & 61.3 & 6.4 & 13.8 & 8.6 & 45.4 & 9.3 \\
\rowcolor{bestrow}\textbf{MemGuard} & \textbf{48.0} & \textbf{7.8} & \textbf{54.9} & \textbf{8.1} & \textbf{44.6} & \textbf{9.8} & \textbf{67.0} & \textbf{5.7} & \textbf{20.7} & \textbf{7.1} & \textbf{50.7} & \textbf{8.1} \\
\midrule
\rowcolor{modelrow}\multicolumn{13}{c}{\textbf{\textit{Gemini-3.1-Pro}}} \\
No Memory & 54.0 & 8.1 & 61.0 & 8.5 & 48.9 & 9.8 & 75.5 & 5.9 & 20.7 & 7.6 & 56.4 & 8.1 \\
Synapse & 55.1 & 8.4 & 62.1 & 8.2 & 50.0 & 9.5 & 76.4 & 6.1 & 20.7 & 7.4 & 57.5 & 8.0 \\
AWM & 57.2 & 7.6 & 64.3 & 7.9 & 52.2 & 9.0 & 79.2 & 5.5 & 17.2 & 6.8 & 59.5 & 7.5 \\
ReasoningBank & 60.4 & 7.1 & 68.1 & 7.3 & 55.6 & 8.3 & 83.0 & 5.0 & 24.1 & 6.4 & 63.1 & 7.0 \\
\rowcolor{bestrow}\textbf{MemGuard} & \textbf{66.3} & \textbf{6.1} & \textbf{74.2} & \textbf{6.3} & \textbf{61.0} & \textbf{7.0} & \textbf{88.7} & \textbf{4.4} & \textbf{34.5} & \textbf{5.1} & \textbf{69.1} & \textbf{5.9} \\
\bottomrule
\end{tabular}
\end{table*}

\begin{table*}[!t]
\centering
\small
\setlength{\tabcolsep}{3.8pt}
\caption{Detailed results on Mind2Web. Results are averaged over five seeds. EA, AF1, SSR, and SR denote element accuracy, action F1, step success rate, and task success rate.}
\label{tab:mind2web-detail}
\begin{tabular}{lcccccccccccc}
\toprule
& \multicolumn{4}{c}{\textbf{Cross-Task}} & \multicolumn{4}{c}{\textbf{Cross-Website}} & \multicolumn{4}{c}{\textbf{Cross-Domain}} \\
\cmidrule(lr){2-5}\cmidrule(lr){6-9}\cmidrule(lr){10-13}
\textbf{Methods} & EA & AF1 & SSR & SR & EA & AF1 & SSR & SR & EA & AF1 & SSR & SR \\
\midrule
\rowcolor{modelrow}\multicolumn{13}{c}{\textbf{\textit{Qwen-3.5-Flash}}} \\
No Memory & 45.2 & 57.8 & 39.0 & 3.0 & 38.6 & 44.0 & 30.5 & 1.5 & 34.9 & 36.8 & 30.2 & 0.8 \\
Synapse & 46.1 & 58.5 & 39.8 & 3.2 & 39.2 & 44.7 & 31.1 & 1.7 & 35.5 & 37.4 & 30.8 & 0.9 \\
AWM & 46.4 & 58.9 & 40.5 & 3.5 & 39.7 & 45.2 & 31.6 & 1.9 & 36.0 & 38.1 & 31.2 & 1.0 \\
ReasoningBank & 50.7 & 60.0 & 43.4 & 4.5 & 42.9 & 50.8 & 33.1 & 2.2 & 39.4 & 40.5 & 35.2 & 1.4 \\
\rowcolor{bestrow}\textbf{MemGuard} & \textbf{54.5} & \textbf{62.2} & \textbf{47.1} & \textbf{6.2} & \textbf{47.2} & \textbf{55.1} & \textbf{36.7} & \textbf{3.8} & \textbf{42.8} & \textbf{43.5} & \textbf{38.4} & \textbf{2.4} \\
\midrule
\rowcolor{modelrow}\multicolumn{13}{c}{\textbf{\textit{Qwen-3.5-Plus}}} \\
No Memory & 54.5 & 64.8 & 48.0 & 4.8 & 47.8 & 55.2 & 39.7 & 4.1 & 42.6 & 44.0 & 39.1 & 1.9 \\
Synapse & 55.1 & 65.5 & 48.6 & 4.9 & 48.6 & 56.0 & 40.5 & 4.2 & 43.2 & 44.8 & 39.7 & 2.0 \\
AWM & 55.8 & 66.0 & 49.0 & 5.2 & 49.4 & 56.8 & 41.0 & 4.4 & 43.8 & 45.6 & 40.1 & 2.0 \\
ReasoningBank & 59.0 & 67.2 & 51.4 & 6.2 & 52.2 & 60.0 & 43.5 & 4.9 & 48.0 & 50.2 & 43.8 & 2.6 \\
\rowcolor{bestrow}\textbf{MemGuard} & \textbf{64.4} & \textbf{71.2} & \textbf{56.8} & \textbf{8.9} & \textbf{59.1} & \textbf{66.2} & \textbf{49.5} & \textbf{7.3} & \textbf{54.4} & \textbf{56.0} & \textbf{49.2} & \textbf{4.2} \\
\midrule
\rowcolor{modelrow}\multicolumn{13}{c}{\textbf{\textit{Gemini-3-Flash}}} \\
No Memory & 48.6 & 59.6 & 42.5 & 3.4 & 40.8 & 48.2 & 33.5 & 2.9 & 37.1 & 37.0 & 34.0 & 1.2 \\
Synapse & 49.4 & 60.3 & 43.0 & 3.5 & 41.5 & 49.1 & 34.0 & 3.0 & 37.8 & 38.7 & 34.7 & 1.3 \\
AWM & 50.0 & 60.8 & 43.5 & 3.7 & 42.6 & 49.4 & 34.7 & 3.1 & 38.4 & 39.5 & 35.2 & 1.4 \\
ReasoningBank & 52.9 & 62.1 & 45.0 & 4.9 & 45.4 & 53.6 & 36.1 & 3.6 & 42.0 & 44.1 & 37.5 & 1.6 \\
\rowcolor{bestrow}\textbf{MemGuard} & \textbf{56.8} & \textbf{64.5} & \textbf{48.8} & \textbf{6.7} & \textbf{50.2} & \textbf{58.6} & \textbf{40.8} & \textbf{5.4} & \textbf{46.2} & \textbf{49.2} & \textbf{41.7} & \textbf{3.0} \\
\midrule
\rowcolor{modelrow}\multicolumn{13}{c}{\textbf{\textit{Gemini-3.1-Pro}}} \\
No Memory & 59.6 & 67.5 & 53.1 & 5.8 & 51.7 & 60.6 & 43.4 & 5.0 & 47.0 & 48.9 & 42.8 & 2.5 \\
Synapse & 60.3 & 68.0 & 53.7 & 5.9 & 52.4 & 61.4 & 44.0 & 5.1 & 47.7 & 49.8 & 43.5 & 2.6 \\
AWM & 61.0 & 68.6 & 54.2 & 6.2 & 53.2 & 62.0 & 44.8 & 5.3 & 48.3 & 50.4 & 43.8 & 2.7 \\
ReasoningBank & 63.8 & 70.1 & 56.0 & 7.3 & 56.5 & 65.0 & 47.2 & 5.9 & 52.3 & 54.2 & 47.4 & 3.1 \\
\rowcolor{bestrow}\textbf{MemGuard} & \textbf{68.1} & \textbf{73.8} & \textbf{60.6} & \textbf{9.3} & \textbf{62.1} & \textbf{70.0} & \textbf{53.0} & \textbf{7.5} & \textbf{58.4} & \textbf{59.0} & \textbf{53.0} & \textbf{4.4} \\
\bottomrule
\end{tabular}
\end{table*}

\section{Prompt Templates}
\label{sec:prompt-templates}

This appendix gives the prompt interfaces used by the verifier, repeated verification views, memory induction, and memory injection. The templates are written as interfaces rather than full benchmark prompts so that the emitted fields can be checked against the schema and governance metadata in Appendix~\ref{sec:implementation-details}.

\subsection{Verifier Prompt}
\label{sec:verifier-prompt}

The verifier receives the task, trajectory, final output, and runtime execution status. It must judge only from logged evidence and must not infer success from the model's confidence. Hidden benchmark labels and final evaluator outcomes are never included in this prompt. The prompt template is:

\begin{quote}
\footnotesize
\raggedright
\textbf{System.} You are an expert evaluator for long-horizon LLM agent trajectories. Judge whether the trajectory provides a reusable experience for future agents. Use only the task, logged trajectory, final output, and runtime-visible status. Do not use hidden benchmark labels, held-out evaluator outcomes, private gold answers, or the model's confidence as evidence.

\textbf{Inputs.} \texttt{task}: \{task\}; \texttt{trajectory}: ordered steps with observations, thoughts, actions, tool outputs, and errors; \texttt{final\_output}: \{output\}; \texttt{runtime\_status}: exit code, public tests, visible web confirmation, tool error, or other status visible to the agent; \texttt{view}: \{full, evidence, risk\}.

\textbf{Criteria.} Score each criterion with one allowed score token from \{1,2,3,4,5\}: (i) \texttt{task\_completion}: whether the user task is completed from visible evidence; (ii) \texttt{evidence\_consistency}: whether the final output is supported by trajectory observations; (iii) \texttt{execution\_validity}: whether commands, patches, web actions, or tool calls are valid and non-spurious; and (iv) \texttt{generalizability}: whether the lesson can transfer beyond the exact instance.

\textbf{Score output.} For each criterion, emit exactly one score token from \{1,2,3,4,5\} before the rationale, so the caller can read the emitted token or its API log probabilities.

\textbf{Label rules.} Return \texttt{verified\_success} only when task completion and evidence consistency are both at least 4 and there is no blocking execution error. Return \texttt{verified\_fail} when visible evidence shows failure, invalid execution, an unsupported final answer, or a harmful strategy. Return \texttt{uncertain} for ambiguous page state, partial tests, missing evidence, conflicting observations, or boundary cases.

\textbf{Output JSON schema.} Return a JSON object with:
\begin{itemize}
\setlength{\itemsep}{0pt}
\setlength{\topsep}{2pt}
\setlength{\leftmargin}{1.1em}
\item \texttt{criteria}: one object per criterion, each with \texttt{score}, \texttt{rationale}, and \texttt{evidence\_span};
\item \texttt{label}: one of \texttt{verified\_success}, \texttt{verified\_fail}, or \texttt{uncertain};
\item \texttt{failure\_guard}: \texttt{NONE} or a short condition to check before reusing the trajectory.
\end{itemize}
The scalar reward and confidence used by later prompts are computed after verification from the criterion scores, score-token distributions, label agreement, and rationale consistency; they are not additional verifier-emitted fields. When the API exposes token log probabilities, MemGuard reads the probabilities assigned to the allowed score tokens and renormalizes over this set. When only the emitted score token is available, MemGuard records the top-1 score and uses a deterministic one-hot distribution for the expectation in Eq.~\ref{eq:reward}.
\end{quote}

\subsection{Verification Views}
\label{sec:verification-views}

Boundary trajectories are re-verified with controlled prompt views over the same log. The base system prompt remains unchanged; only the \texttt{view} instruction is changed:

\begin{quote}
\footnotesize
\raggedright
\textbf{Full view.} Evaluate the complete trajectory. Consider all observations, actions, tool outputs, final output, and runtime-visible status.

\textbf{Evidence view.} Ignore unrelated intermediate exploration. Focus on the minimal spans that directly support or contradict the final output. If no span supports the final output, lower \texttt{evidence\_consistency} and explain the missing evidence.

\textbf{Risk view.} Prioritize failure modes: wrong entity or row selection, stale page state, invalid command, failing test, unsupported assumption, loop, hallucinated file path, or task-specific shortcut that may not transfer. If a memory could cause future harm when retrieved, emit a \texttt{failure\_guard}.
\end{quote}

\subsection{Memory Induction Prompt}
\label{sec:induction-prompt}

The memory induction prompt converts a verifier-processed trajectory into auditable memory records:

\begin{quote}
\footnotesize
\raggedright
\textbf{System.} You extract reusable memories from verifier-processed agent trajectories. Produce at most three candidate records. Each retained record must cite an evidence span and must be useful beyond the current task. Do not store exact benchmark answers, private labels, raw credentials, hidden evaluator feedback, or brittle selectors such as ``click the first row'' unless the memory is a failure guard that warns against the shortcut.

\textbf{Inputs.} \texttt{task}; \texttt{trajectory}; \texttt{final\_output}; \texttt{runtime\_status}; verifier output, including \texttt{criteria} entries with \texttt{rationale} and \texttt{evidence\_span}, plus \texttt{failure\_guard}; \texttt{verifier\_label}; \texttt{reward} computed from criterion scores via Eq.~\ref{eq:reward}; \texttt{confidence}; and retrieved memories that influenced the run.

\textbf{Rules.} Use \texttt{procedural\_hint} for reusable reasoning or debugging procedure, \texttt{tool\_usage} for command or environment interaction patterns, and \texttt{failure\_avoidance} for negative lessons. A failure memory must be phrased as a condition or check, not as an action recipe. Reject a candidate if its evidence span is missing, if it merely restates the task, or if it depends on a one-off object name.

\textbf{Output JSON schema.} Return \texttt{\{"records":[...]\}}. Each record contains:
\begin{itemize}
\setlength{\itemsep}{0pt}
\setlength{\topsep}{2pt}
\setlength{\leftmargin}{1.1em}
\item required fields: \texttt{type}, \texttt{title}, \texttt{description}, \texttt{content}, \texttt{applicability}, \texttt{risk}, \texttt{guard\_condition}, \texttt{evidence\_span}, and \texttt{reject\_reason};
\item \texttt{type}: \texttt{procedural\_hint}, \texttt{tool\_usage}, or \texttt{failure\_avoidance};
\item \texttt{risk}: \texttt{none}, \texttt{low}, \texttt{medium}, or \texttt{high};
\item \texttt{guard\_condition}: \texttt{NONE} for positive records or a short check condition for failure-avoidance records, derived from the verifier guard when applicable;
\item \texttt{evidence\_span}: a trajectory span or \texttt{NONE};
\item \texttt{reject\_reason}: \texttt{NONE} for retained records or a short rejection reason.
\end{itemize}
The governance layer then normalizes \texttt{guard\_condition} into the stored failure-guard field and attaches \texttt{source\_status}, \texttt{quality\_score}, \texttt{verifier\_label}, reward, confidence, verification time, and the initial lifecycle state. Here \texttt{source\_status} records the runtime-visible execution status category (for example, public-test pass, public-test fail, visible web confirmation, tool error, or unknown), while \texttt{verifier\_label} records the verifier's judgment. This keeps prompt-based induction separate from admission decisions.
\end{quote}

\subsection{Memory Injection Prompt}
\label{sec:memory-injection-prompt}

At test time, retrieved memory is rendered as a compact block before the current task. Positive memories and failure guards are separated so that the agent can distinguish suggested strategies from constraints:

\begin{quote}
\footnotesize
\raggedright
\textbf{System memory block.} The following governed memories were derived from previous tasks and may help only if their applicability condition matches the current task. Before using a memory, check whether its condition holds. Do not treat memory as ground truth, do not follow it when the current observation contradicts it, and do not copy task-specific names, paths, rows, DOM identifiers, or answers unless they are present in the current task.

\textbf{Positive memories.} Up to five active procedural or tool-usage records, rendered as \texttt{[id] title; content; applicability; confidence}.

\textbf{Summary memories.} Up to two compressed records when the active bank exceeds the memory budget, rendered as \texttt{[id] summary; covered record ids; applicability}.

\textbf{Failure guards.} Up to two high-confidence negative records, rendered as \texttt{[id] risk; check-before-acting; evidence summary}. Treat these as constraints. If a guard applies, avoid the risky action unless current evidence resolves the risk.
\end{quote}

\subsection{Prompt Example}
\label{sec:prompt-example}

The following abbreviated example illustrates how a misleading web trajectory becomes a guard rather than a positive recipe.

\begin{quote}
\footnotesize
\raggedright
\textbf{Input.} Task: update the target user in an admin table. Trajectory excerpt: after filtering, the agent clicked the first visible row and saved the edit. Runtime status: the visible confirmation appeared, but the verifier risk view notes that the row identity was never checked after sorting.

\textbf{Verifier output excerpt.} The verifier returns \texttt{uncertain}, with criterion scores 3, 3, 3, and 2 for task completion, evidence consistency, execution validity, and generalizability. The confirmation is visible, but the evidence does not establish the edited row's identity. It emits the guard: before editing a table row, verify the row identity using a stable field after filtering or sorting.

\textbf{Governed memory.} Type: \texttt{failure\_avoidance}. Title: verify table-row identity after sorting. Content: do not assume the first visible row is the target after filtering or sorting; check a stable identifier before editing or submitting changes. Applicability: admin tables, search results, issue lists, or any sortable table. Risk: \texttt{high}. Initial state: \texttt{provisional}.
\end{quote}

\section{Additional Experimental Analyses}
\label{sec:efficiency-ablation}

This section collects result-oriented diagnostics that extend the main experiments: long-horizon accumulation, deployment cost, verifier reliability, sensitivity checks, control variants, and component ablations. Implementation parameters and audit protocols are separated into Appendix~\ref{sec:implementation-details}, while memory-bank health and failure-memory behavior are analyzed in Appendix~\ref{sec:robustness-governance}.

\subsection{Long-Horizon Task-Stream Accumulation}
\label{sec:accumulation}

Long-running agents face continuous task streams, so an effective memory system should become more useful as experience accumulates. These curves are not independent per-task evaluations: within each benchmark-model run, one memory bank grows across the evaluated task order and is not reset between tasks. We divide each benchmark stream into five checkpoints at 20\%, 40\%, 60\%, 80\%, and 100\%, and compute cumulative average performance at each checkpoint. The curves cover Qwen-3.5-Flash, Qwen-3.5-Plus, Gemini-3-Flash, and Gemini-3.1-Pro. Each 100\% checkpoint matches the corresponding final value in Table~\ref{tab:terminal-swe} and the detailed web tables in Tables~\ref{tab:webarena-detail} and~\ref{tab:mind2web-detail}. For Mind2Web, the success curve plots average SSR across Cross-Task, Cross-Website, and Cross-Domain.

\begin{figure*}[!t]
\centering
\includegraphics[width=0.98\textwidth]{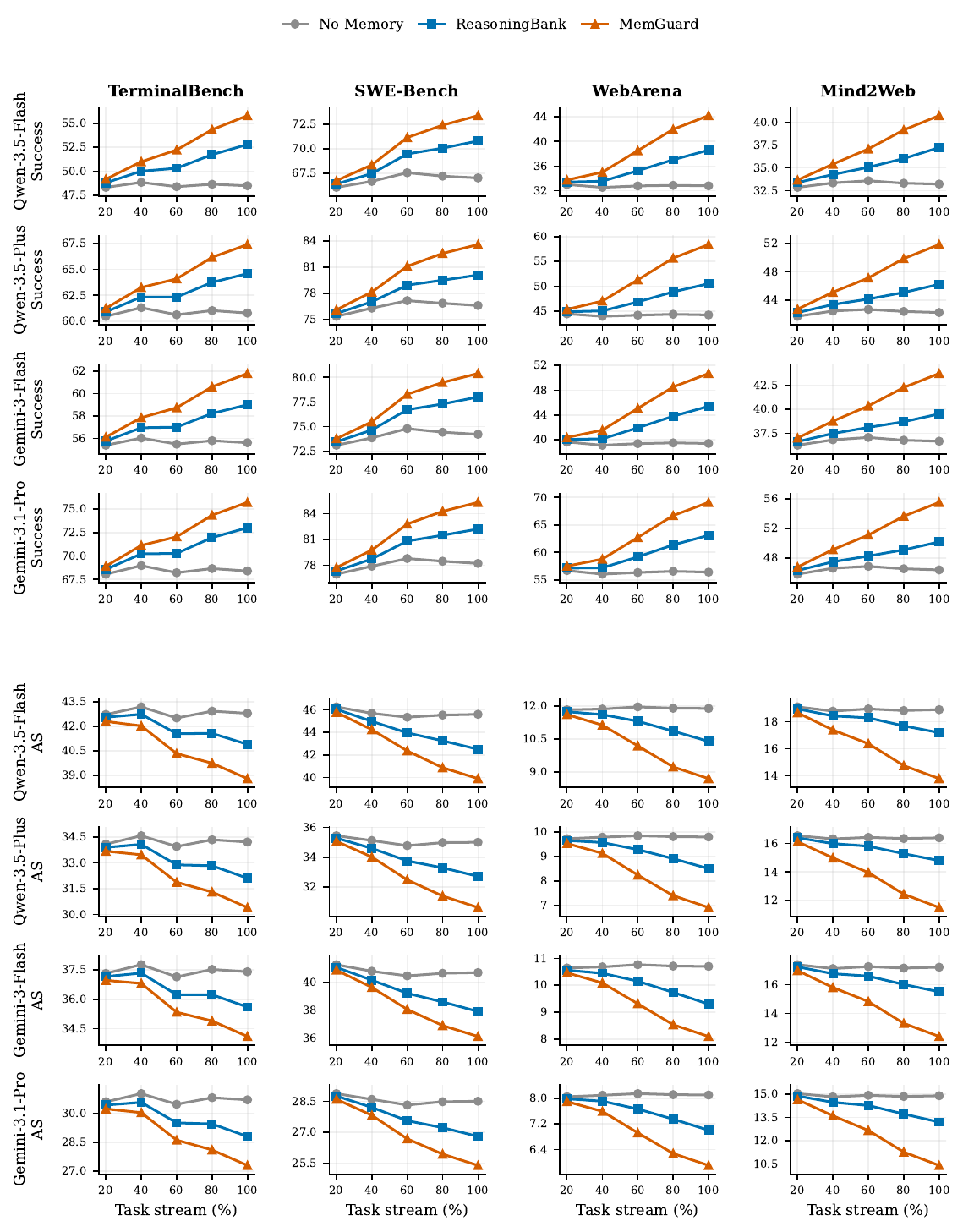}
\caption{Task-stream accumulation diagnostics. The top four rows report cumulative success metrics; the bottom four rows report cumulative AS, where lower is better. Each subplot shares the same task-stream checkpoints. For Mind2Web, the success metric is average SSR across Cross-Task, Cross-Website, and Cross-Domain. MemGuard's late-stream gains coincide with lower AS rather than longer trajectories.}
\label{fig:accumulation}
\end{figure*}

Figure~\ref{fig:accumulation} shows that the methods are close early in the task stream, when few reusable memories exist and cumulative averages are sensitive to task difficulty. As more tasks are processed, ReasoningBank gradually exceeds No Memory, and MemGuard separates more clearly in later checkpoints, especially on WebArena and Mind2Web.

The AS rows show the analogous efficiency trend. No Memory mainly reflects natural task difficulty variation, while ReasoningBank reduces some late-stage exploration. MemGuard further reduces AS as the task stream grows, suggesting that its gains are not only from solving more tasks but also from avoiding repeated failed actions and locating useful actions earlier. This paired pattern is important: the success curves improve while the AS curves fall, so the late-stream gains are not obtained by simply taking longer trajectories.

\subsection{Extended SWE-bench Full Stream}
\label{sec:swe-full-long-horizon}

To test memory governance beyond the main benchmark horizons, we run Qwen-3.5-Plus on all 2,294 unique SWE-bench Full issues as one chronological, non-repeating stream. This extends the 500-issue Verified stream by $4.6\times$; Verified issues remain once at their chronological positions because they are a subset of Full. Tasks are ordered by issue creation time, with ties broken by instance ID. The repository workspace, tools, and conversation reset for every issue, while the memory bank persists across the entire stream. ReasoningBank and MemGuard use the same task order, 60-step limit, decoding settings, retrieval budget, active-memory budget, and two matched seeds. Gold patches, evaluator-only tests, and post-hoc outcomes are unavailable to every online component.

\begin{table*}[t]
\centering
\small
\setlength{\tabcolsep}{5pt}
\caption{Chronological continual-stream evaluation on all 2,294 SWE-bench Full issues. Results report mean $\pm$ sample standard deviation over two matched seeds. Checkpoint gaps are MemGuard minus ReasoningBank for Resolve Rate and AS.}
\label{tab:swe-full-long-horizon}
\begin{tabular}{lcccccc}
\toprule
\textbf{Method} & \textbf{Resolve Rate} & \textbf{AS} & \textbf{20\% gap} & \textbf{40\% gap} & \textbf{60\% gap} & \textbf{80\% gap} \\
\midrule
ReasoningBank & $49.22{\pm}1.17$ & $43.25{\pm}0.49$ & \multicolumn{4}{c}{\multirow{2}{*}{RR: $0.00, +0.33, +1.56, +2.24$; AS: $-0.25, -0.45, -0.70, -1.30$}} \\
\textbf{MemGuard} & $\mathbf{52.18{\pm}0.80}$ & $\mathbf{41.15{\pm}1.06}$ & \multicolumn{4}{c}{} \\
\bottomrule
\end{tabular}
\end{table*}

At the full-stream checkpoint, MemGuard gains 2.96 Resolve Rate points and reduces AS by 2.10 steps. Its cumulative Resolve Rate gap grows from 0.00 at 20\% to 2.96 points at 100\%; in non-overlapping windows, the gaps are 0.00, 0.66, 4.04, 4.25, and 5.88 points, while AS is lower in every window. The widening late-stream separation is consistent with lifecycle governance becoming more useful as persistent memory accumulates and approaches its fixed 384-record capacity.

\subsection{Verifier-Only Control and Deployment Cost}
\label{sec:verifier-control-details}

\begin{table*}[!t]
\centering
\small
\setlength{\tabcolsep}{3.2pt}
\renewcommand{\arraystretch}{1.08}
\caption{Deployment cost accounting across all four benchmarks.}
\label{tab:overhead}
\begin{tabular*}{\textwidth}{@{\extracolsep{\fill}}lcccccc}
\toprule
{\normalsize\textbf{Method}} & {\normalsize\textbf{AS}} & {\normalsize\textbf{Agent tok./task}} & {\normalsize\textbf{Extra tok./task}} & {\normalsize\textbf{Total tok./task}} & {\normalsize\textbf{Latency/task}} & {\normalsize\textbf{Storage MB}} \\
\midrule
\rowcolor{modelrow}\multicolumn{7}{c}{\normalsize\textbf{\textit{Qwen-3.5-Flash}}} \\
No Memory & 29.8 & 64.0k & 0.0k & 64.0k & 103.2s & 0.0 \\
Synapse & 29.7 & 66.4k & 0.0k & 66.4k & 111.6s & 1.3 \\
AWM & 28.9 & 67.0k & 0.0k & 67.0k & 112.8s & 1.6 \\
ReasoningBank & 27.8 & 68.4k & 0.0k & 68.4k & 115.4s & 2.5 \\
Verifier-only Filter & 26.7 & 62.8k & 2.6k & 65.4k & 110.2s & 2.4 \\
\rowcolor{bestrow}\textbf{MemGuard} & \textbf{25.3} & \textbf{61.0k} & \textbf{4.0k} & \textbf{65.0k} & \textbf{108.9s} & \textbf{3.6} \\
\midrule
\rowcolor{modelrow}\multicolumn{7}{c}{\normalsize\textbf{\textit{Qwen-3.5-Plus}}} \\
No Memory & 23.9 & 78.5k & 0.0k & 78.5k & 164.2s & 0.0 \\
Synapse & 24.1 & 82.0k & 0.0k & 82.0k & 176.3s & 1.2 \\
AWM & 23.5 & 83.0k & 0.0k & 83.0k & 179.3s & 1.5 \\
ReasoningBank & 22.0 & 84.0k & 0.0k & 84.0k & 184.0s & 2.1 \\
Verifier-only Filter & 21.1 & 78.8k & 2.9k & 81.7k & 181.0s & 2.0 \\
\rowcolor{bestrow}\textbf{MemGuard} & \textbf{19.9} & \textbf{76.5k} & \textbf{4.5k} & \textbf{81.0k} & \textbf{174.8s} & \textbf{3.0} \\
\midrule
\rowcolor{modelrow}\multicolumn{7}{c}{\normalsize\textbf{\textit{Gemini-3-Flash}}} \\
No Memory & 26.5 & 71.5k & 0.0k & 71.5k & 101.8s & 0.0 \\
Synapse & 26.3 & 75.6k & 0.0k & 75.6k & 108.9s & 1.2 \\
AWM & 25.9 & 76.4k & 0.0k & 76.4k & 110.8s & 1.5 \\
ReasoningBank & 24.6 & 77.0k & 0.0k & 77.0k & 114.0s & 2.3 \\
Verifier-only Filter & 23.8 & 70.6k & 3.4k & 74.0k & 109.5s & 2.2 \\
\rowcolor{bestrow}\textbf{MemGuard} & \textbf{22.7} & \textbf{67.9k} & \textbf{4.9k} & \textbf{72.8k} & \textbf{106.9s} & \textbf{3.3} \\
\midrule
\rowcolor{modelrow}\multicolumn{7}{c}{\normalsize\textbf{\textit{Gemini-3.1-Pro}}} \\
No Memory & 20.6 & 130.5k & 0.0k & 130.5k & 329.0s & 0.0 \\
Synapse & 20.8 & 138.8k & 0.0k & 138.8k & 354.0s & 1.1 \\
AWM & 20.0 & 140.0k & 0.0k & 140.0k & 357.0s & 1.4 \\
ReasoningBank & 19.0 & 141.2k & 0.0k & 141.2k & 363.0s & 2.0 \\
Verifier-only Filter & 18.1 & 131.1k & 3.4k & 134.5k & 350.0s & 1.9 \\
\rowcolor{bestrow}\textbf{MemGuard} & \textbf{17.3} & \textbf{127.0k} & \textbf{5.8k} & \textbf{132.8k} & \textbf{344.0s} & \textbf{2.9} \\
\bottomrule
\end{tabular*}
\vspace{2pt}
\begin{minipage}{\textwidth}
\raggedright
\footnotesize\emph{Note.} AS is the unweighted macro-average over the four benchmark settings in Table~\ref{tab:terminal-swe}. Agent tokens include task execution and compact memory-injection prompts; extra tokens count verifier, re-verification, induction, and governance calls outside AS. No Memory is a no-retrieval, no-verification lower-cost reference rather than an accuracy--cost optimum. Storage is the final serialized memory-bank footprint.
\end{minipage}
\end{table*}

The Verifier-only Filter rows in Table~\ref{tab:terminal-swe} separate two effects that would otherwise be confounded. Verifier-only Filter tests whether a trajectory verifier alone is enough; MemGuard tests whether verifier outputs must persist as memory metadata and feed back into retrieval and governance. The consistent gap between the two variants shows that one-time filtering captures only part of the gain.

Table~\ref{tab:overhead} reports one deployment-cost accounting convention: agent tokens include task execution and compact memory injection, while extra tokens count verifier, re-verification, induction, and governance calls outside AS. Memory methods incur extra memory-processing cost relative to No Memory. MemGuard also adds verifier and governance calls, but its lower AS keeps total tokens and latency below the other memory and verifier baselines while remaining above the no-memory lower bound.

\subsection{Verifier Reliability and Label Noise}
\label{sec:verifier-reliability}

\begin{table}[!htbp]
\centering
\small
\setlength{\tabcolsep}{2.0pt}
\caption{Verifier calibration audit on 800 sampled Qwen trajectories. Values are cases; FA/FR denote false accepts/rejects, and blocked FA denotes false accepts not promoted to active positive memory.}
\label{tab:verifier-calibration}
\resizebox{\columnwidth}{!}{%
\begin{tabular}{llccccc}
\toprule
\textbf{Backbone} & \textbf{Benchmark} & $n$ & Agree & FA & FR & Blocked FA \\
\midrule
\multirow{5}{*}{Qwen-3.5-Plus}
& Terminal-Bench 2.0 & 100 & 86 & 6 & 8 & 4 \\
& SWE-Bench Verified & 100 & 85 & 7 & 8 & 5 \\
& WebArena & 100 & 87 & 5 & 8 & 3 \\
& Mind2Web & 100 & 86 & 8 & 6 & 6 \\
& \textbf{Overall} & \textbf{400} & \textbf{344} & \textbf{26} & \textbf{30} & \textbf{18} \\
\midrule
\multirow{5}{*}{Qwen-3.5-Flash}
& Terminal-Bench 2.0 & 100 & 82 & 9 & 9 & 5 \\
& SWE-Bench Verified & 100 & 81 & 10 & 9 & 6 \\
& WebArena & 100 & 84 & 8 & 8 & 4 \\
& Mind2Web & 100 & 85 & 7 & 8 & 5 \\
& \textbf{Overall} & \textbf{400} & \textbf{332} & \textbf{34} & \textbf{34} & \textbf{20} \\
\bottomrule
\end{tabular}
}
\end{table}

\paragraph{Independent benchmark-outcome calibration.}
The author audit above covers all four verifier criteria. We separately validate the frozen trajectory-level task-completion decision against post-hoc benchmark outcomes that were never available to the agent, verifier, induction module, or admission policy. We join 51,896 of 52,280 task-seed trajectories (99.3\%) to official Terminal-Bench 2.0 success, SWE-Bench Verified resolved status, official WebArena outcomes, and strict Mind2Web task success derived from the benchmark gold action sequence. With the prompts, score mapping, aggregation, and threshold frozen before this join, the task-completion decision obtains 88.1\% macro balanced accuracy and a 9.3\% false-positive rate. This check provides an outcome-based test independent of author inspection, while the smaller audit remains necessary for criteria that do not reduce to task success.

Because verifier errors can directly affect memory quality, we audit balanced samples of completed trajectories under both Qwen-3.5-Plus and Qwen-3.5-Flash. Table~\ref{tab:verifier-calibration} reports cases rather than percentages. The weaker Qwen-3.5-Flash verifier has lower agreement and more false accepts, but 38 of 60 observed false accepts are blocked by governance because their confidence, evidence consistency, or conflict state prevents promotion to active positive memory. This pattern is important for memory governance: false rejections may discard useful experience, but false accepts can turn harmful experience into active memory. MemGuard therefore uses stricter activation thresholds, repeated verification for boundary cases, and provisional states for low-confidence records.

\begin{table*}[t]
\centering
\small
\setlength{\tabcolsep}{4pt}
\caption{Downstream tracing of verifier false accepts. We trace false-accept records from the calibration audit through governance, retrieval, action influence, and downstream task failure. Counts are records, not percentages.}
\label{tab:false-accept-tracing}
\begin{tabular}{llrrrrrr}
\toprule
\textbf{Backbone} & \textbf{Benchmark} & FA & Blocked & Active & Retrieved & Influenced & Failure \\
\midrule
\multirow{4}{*}{Qwen-3.5-Plus}
& Terminal-Bench 2.0 & 6 & 4 & 2 & 1 & 1 & 0 \\
& SWE-Bench Verified & 7 & 5 & 2 & 2 & 1 & 1 \\
& WebArena & 5 & 3 & 2 & 2 & 1 & 0 \\
& Mind2Web & 8 & 6 & 2 & 1 & 1 & 0 \\
\midrule
\multirow{4}{*}{Qwen-3.5-Flash}
& Terminal-Bench 2.0 & 9 & 5 & 4 & 3 & 2 & 1 \\
& SWE-Bench Verified & 10 & 6 & 4 & 3 & 2 & 1 \\
& WebArena & 8 & 4 & 4 & 4 & 3 & 2 \\
& Mind2Web & 7 & 5 & 2 & 2 & 1 & 1 \\
\midrule
\textbf{Overall} & \textbf{All} & \textbf{60} & \textbf{38} & \textbf{22} & \textbf{18} & \textbf{12} & \textbf{6} \\
\bottomrule
\end{tabular}
\end{table*}

Table~\ref{tab:false-accept-tracing} follows the same false accepts into later memory use. Most are blocked before becoming active positive memories; among the 22 active false accepts, 18 are later retrieved, 12 influence an agent action, and 6 are associated with an observed downstream task failure. This tracing does not eliminate verifier dependence, but it bounds the practical failure pathway in the audited stream and shows why false accepts are more serious than false rejects for long-term memory systems.

\paragraph{Stylized risk accounting.}
The audit can be interpreted through a simple event decomposition. Let $H_m$ denote that candidate memory $m$ is harmful, $F_m$ that the verifier falsely accepts it, $A_m$ that it remains active after admission governance, and $I_{m,t}$ that it is injected at a future retrieval opportunity $t$. A verifier-only filter uses the verifier decision as the final admission decision, so
\[
\Pr(A_m \mid H_m) = \Pr(F_m \mid H_m) = \epsilon .
\]
MemGuard keeps the same verifier but inserts additional gates after a false accept. Define
\[
\begin{aligned}
\rho_{\mathrm{adm}}&=\Pr(A_m\mid F_m,H_m),\\
\rho_{\mathrm{inj}}&=\Pr(I_{m,t}\mid A_m,F_m,H_m).
\end{aligned}
\]
Then the probability that a harmful memory is both falsely accepted and later injected at opportunity $t$ factors as
\[
\Pr(I_{m,t},A_m,F_m\mid H_m)
=\epsilon\,\rho_{\mathrm{adm}}\,\rho_{\mathrm{inj}} .
\]
Under a verifier-only filter with the same retrieval budget, the corresponding term is $\epsilon\rho_{\mathrm{base}}$, where $\rho_{\mathrm{base}}$ is the retrieval probability for an accepted harmful memory. Governance therefore helps when $\rho_{\mathrm{adm}}\rho_{\mathrm{inj}}<\rho_{\mathrm{base}}$: the same false verifier label is less likely to reach injection. The decomposition formalizes the failure pathway measured in Table~\ref{tab:false-accept-tracing}.

\begin{table*}[!htbp]
\centering
\scriptsize
\setlength{\tabcolsep}{3pt}
\caption{Sensitivity to simulated verifier label noise with a verifier-only control. Each cell reports success metric / AS: SR for Terminal-Bench 2.0 and WebArena, Resolve Rate for SWE-Bench Verified, and SSR for Mind2Web. Label noise randomly corrupts verifier labels before memory admission.}
\label{tab:verifier-noise}
\resizebox{\textwidth}{!}{%
\begin{tabular}{lllcccc}
\toprule
\textbf{Model} & \textbf{Method} & \textbf{Noise}
& \textbf{Terminal-Bench 2.0}
& \textbf{SWE-Bench Verified}
& \textbf{WebArena}
& \textbf{Mind2Web} \\
\midrule
\multirow{8}{*}{Qwen-3.5-Flash}
& Verifier-only & 0\% & 53.0 / 40.7 & 71.9 / 41.5 & 42.6 / 9.1 & 37.8 / 15.4 \\
& Verifier-only & 10\% & 52.1 / 42.6 & 70.7 / 42.9 & 40.1 / 10.2 & 37.2 / 15.9 \\
& Verifier-only & 20\% & 50.6 / 44.0 & 69.4 / 45.1 & 38.2 / 10.8 & 36.1 / 16.8 \\
& Verifier-only & 30\% & 49.8 / 45.2 & 67.9 / 46.0 & 35.6 / 11.8 & 34.9 / 17.1 \\
\cmidrule(lr){2-7}
& MemGuard & 0\% & 55.8 / 38.8 & 73.4 / 39.9 & 44.2 / 8.7 & 40.7 / 13.8 \\
& MemGuard & 10\% & 54.7 / 40.0 & 72.5 / 41.1 & 42.7 / 9.4 & 40.3 / 14.1 \\
& MemGuard & 20\% & 53.9 / 39.6 & 71.0 / 42.8 & 41.0 / 9.1 & 39.4 / 15.0 \\
& MemGuard & 30\% & 52.9 / 42.3 & 70.1 / 43.5 & 39.8 / 10.6 & 38.6 / 15.3 \\
\midrule
\multirow{8}{*}{Qwen-3.5-Plus}
& Verifier-only & 0\% & 64.8 / 32.4 & 81.9 / 31.6 & 52.0 / 8.4 & 50.0 / 11.9 \\
& Verifier-only & 10\% & 63.4 / 34.2 & 80.2 / 33.5 & 49.8 / 9.2 & 49.1 / 12.4 \\
& Verifier-only & 20\% & 61.0 / 35.8 & 78.1 / 35.0 & 46.4 / 10.1 & 47.5 / 13.1 \\
& Verifier-only & 30\% & 59.5 / 37.9 & 76.0 / 36.2 & 43.5 / 10.9 & 46.0 / 13.8 \\
\cmidrule(lr){2-7}
& MemGuard & 0\% & 67.4 / 30.4 & 83.6 / 30.6 & 58.4 / 6.9 & 51.8 / 11.5 \\
& MemGuard & 10\% & 66.1 / 31.7 & 82.8 / 31.5 & 56.8 / 7.2 & 51.4 / 11.9 \\
& MemGuard & 20\% & 65.0 / 31.2 & 81.2 / 33.0 & 54.0 / 7.8 & 50.5 / 11.7 \\
& MemGuard & 30\% & 63.5 / 33.8 & 79.4 / 33.4 & 52.1 / 8.5 & 49.4 / 12.9 \\
\bottomrule
\end{tabular}
}
\end{table*}

We simulate verifier mistakes by randomly corrupting verifier labels before admission while keeping retrieval, memory budgets, and task order fixed. Table~\ref{tab:verifier-noise} compares MemGuard with Verifier-only Filter under the same corrupted verifier labels. The trend is intentionally benchmark-dependent: binary success metrics on Terminal-Bench 2.0 and WebArena degrade more than Mind2Web's step-level SSR, and AS changes are non-linear because task-level step counts are discrete and order-sensitive. The expected mechanism is governance buffering: noisy verifier labels can still create bad admissions, but provisional states, confidence-aware retrieval, conflict checks, and failure-guard separation reduce the chance that an erroneous label is injected as an active positive memory. Across the eight matched benchmark--backbone comparisons at 30\% noise, MemGuard loses 0.3--2.6 fewer metric points from the clean-verifier setting than the verifier-only control, supporting the claim that persistent verifier signals plus lifecycle governance buffer label errors that one-time filtering cannot.

\subsection{Score-Token Granularity}
\label{sec:score-granularity}

We vary the number of verifier score tokens while keeping the same criteria, prompts, and admission thresholds. Table~\ref{tab:score-granularity} reports Qwen-3.5-Plus on WebArena and Mind2Web as a diagnostic subset, where verifier uncertainty is most frequent. A 3-token scale is cheaper but loses useful boundary information. A 10-token scale gives little additional benefit over the default 5-token scale while increasing verifier output length.

\begin{table}[!htbp]
\centering
\small
\setlength{\tabcolsep}{4pt}
\caption{Sensitivity to score-token granularity on Qwen-3.5-Plus.}
\label{tab:score-granularity}
\begin{tabular}{lcccc}
\toprule
\multirow{2}{*}{$G$}
& \multicolumn{2}{c}{\textbf{WebArena}}
& \multicolumn{2}{c}{\textbf{Mind2Web}} \\
\cmidrule(lr){2-3}\cmidrule(lr){4-5}
& SR $\uparrow$ & AS $\downarrow$ & SSR $\uparrow$ & AS $\downarrow$ \\
\midrule
3 & 57.1 & 7.1 & 50.7 & 11.8 \\
\textbf{5} & \textbf{58.4} & \textbf{6.9} & \textbf{51.8} & \textbf{11.5} \\
10 & 58.1 & 7.0 & 52.0 & 11.5 \\
\bottomrule
\end{tabular}
\end{table}

\subsection{Descriptor Simplification}
\label{sec:descriptor-simplification}

The score-token granularity analysis retains the full lifecycle descriptor, so it does not isolate the downstream value of reward and uncertainty beyond label and confidence. We therefore compare full MemGuard with a simplified decision descriptor $d'_m=(\ell_m,c_m)$. The simplified variant keeps the verifier prompt and confidence computation unchanged, but removes continuous reward and uncertainty as separate inputs to admission, retrieval ranking, and conflict resolution. All task orders, memory and retrieval budgets, decoding settings, and hard filters remain matched.

\begin{table*}[t]
\centering
\small
\setlength{\tabcolsep}{4pt}
\caption{Descriptor simplification on Qwen-3.5-Plus. Each cell reports success metric / AS as mean $\pm$ standard deviation over five seeds.}
\label{tab:descriptor-simplification}
\begin{tabular}{lcccc}
\toprule
\textbf{Variant} & \textbf{Terminal-Bench 2.0} & \textbf{SWE-Bench Verified} & \textbf{WebArena} & \textbf{Mind2Web} \\
\midrule
\textbf{Full MemGuard} & $67.4{\pm}1.9/30.4{\pm}1.7$ & $83.6{\pm}1.6/30.6{\pm}1.2$ & $58.4{\pm}0.8/6.9{\pm}0.3$ & $51.8{\pm}0.5/11.5{\pm}0.4$ \\
$\{\text{label},\text{confidence}\}$ & $66.3{\pm}1.5/31.0{\pm}2.2$ & $82.9{\pm}1.7/30.9{\pm}1.1$ & $56.9{\pm}1.3/7.3{\pm}0.6$ & $51.1{\pm}0.5/11.7{\pm}0.5$ \\
\bottomrule
\end{tabular}
\end{table*}

The simplified descriptor retains most of the gain and remains above Verifier-only Filter on every primary success metric. The full descriptor improves the four success metrics by 0.7--1.5 points and reduces AS by 0.2--0.6 steps. We interpret reward and uncertainty as modest complementary signals for downstream ranking and boundary cases rather than the main source of the gain.

\subsection{Admission Threshold and View Sensitivity}
\label{sec:threshold-sensitivity}

We next test whether the main results depend on a narrow choice of admission thresholds. Table~\ref{tab:threshold-sensitivity} varies the activation thresholds and uncertainty mixture on Qwen-3.5-Plus and Qwen-3.5-Flash for the two benchmarks with the highest boundary-case rates. This subset is intended as a stress test of the admission mechanism rather than a full model sweep. The default setting is strongest in this subset, but nearby settings remain close, suggesting that MemGuard's gains are not produced by a brittle threshold choice.

\begin{table}[!htbp]
\centering
\footnotesize
\setlength{\tabcolsep}{2pt}
\renewcommand{\arraystretch}{0.96}
\caption{Admission-threshold sensitivity on Qwen backbones.}
\label{tab:threshold-sensitivity}
\resizebox{\columnwidth}{!}{%
\begin{tabular}{llcccc}
\toprule
\multirow{2}{*}{\textbf{Backbone}} & \multirow{2}{*}{\textbf{Setting}}
& \multicolumn{2}{c}{\textbf{WebArena}}
& \multicolumn{2}{c}{\textbf{Mind2Web}} \\
\cmidrule(lr){3-4}\cmidrule(lr){5-6}
& & SR & AS & SSR & AS \\
\midrule
Qwen-3.5-Plus & $R{\geq}.65,c{\geq}.55,\alpha{=}.5$ & 57.5 & 7.1 & 50.9 & 11.8 \\
\textbf{Qwen-3.5-Plus} & \boldmath\textbf{$R{\geq}.70,c{\geq}.60,\alpha{=}.5$}\unboldmath & \textbf{58.4} & \textbf{6.9} & \textbf{51.8} & \textbf{11.5} \\
Qwen-3.5-Plus & $R{\geq}.75,c{\geq}.65,\alpha{=}.5$ & 57.8 & 7.0 & 51.1 & 11.7 \\
Qwen-3.5-Plus & $R{\geq}.70,c{\geq}.60,\alpha{=}.3$ & 57.6 & 7.0 & 51.3 & 11.7 \\
Qwen-3.5-Plus & $R{\geq}.70,c{\geq}.60,\alpha{=}.7$ & 58.0 & 7.0 & 51.5 & 11.6 \\
\midrule
Qwen-3.5-Flash & $R{\geq}.65,c{\geq}.55,\alpha{=}.5$ & 43.3 & 8.9 & 39.7 & 14.1 \\
\textbf{Qwen-3.5-Flash} & \boldmath\textbf{$R{\geq}.70,c{\geq}.60,\alpha{=}.5$}\unboldmath & \textbf{44.2} & \textbf{8.7} & \textbf{40.7} & \textbf{13.8} \\
Qwen-3.5-Flash & $R{\geq}.75,c{\geq}.65,\alpha{=}.5$ & 43.6 & 8.9 & 40.0 & 14.0 \\
Qwen-3.5-Flash & $R{\geq}.70,c{\geq}.60,\alpha{=}.3$ & 43.5 & 8.9 & 40.2 & 14.0 \\
Qwen-3.5-Flash & $R{\geq}.70,c{\geq}.60,\alpha{=}.7$ & 43.9 & 8.8 & 40.4 & 13.9 \\
\bottomrule
\end{tabular}
}
\renewcommand{\arraystretch}{1.0}
\end{table}

The repeated views are not independent verifiers; they are controlled prompt views over the same logged trajectory. Table~\ref{tab:view-agreement} therefore reports view agreement as a diagnostic rather than as ensemble accuracy. Web tasks produce lower agreement because page state, element identity, and unsupported assumptions are more often ambiguous, which explains why repeated verification is triggered more frequently on WebArena and Mind2Web. The additional Qwen-3.5-Flash block tests whether this pattern persists under a weaker backbone.

\begin{table*}[!htbp]
\centering
\footnotesize
\setlength{\tabcolsep}{5pt}
\caption{Agreement among verification views on Qwen backbones. Rates are percentages of completed trajectories.}
\label{tab:view-agreement}
\begin{tabular}{llcccc}
\toprule
\textbf{Backbone} & \textbf{Benchmark} & Boundary & Full/Evid. & Full/Risk & All-view \\
& & rate & agree & agree & agree \\
\midrule
\multirow{4}{*}{Qwen-3.5-Flash}
& Terminal-Bench 2.0 & 15.8 & 87.4 & 82.6 & 79.8 \\
& SWE-Bench Verified & 18.1 & 85.2 & 80.9 & 78.1 \\
& WebArena & 24.3 & 81.6 & 75.9 & 71.3 \\
& Mind2Web & 26.8 & 80.1 & 74.2 & 69.8 \\
\midrule
\multirow{4}{*}{Qwen-3.5-Plus}
& Terminal-Bench 2.0 & 14.2 & 89.6 & 85.1 & 82.4 \\
& SWE-Bench Verified & 16.5 & 87.8 & 83.6 & 80.7 \\
& WebArena & 21.7 & 84.2 & 78.5 & 74.1 \\
& Mind2Web & 23.9 & 82.7 & 76.8 & 72.6 \\
\bottomrule
\end{tabular}
\end{table*}

\subsection{Runtime Feedback, Weight, and Cross-Family Verifier Results}
\label{sec:feedback-verifier-controls}

This subsection groups controls that test whether MemGuard's gains depend on privileged runtime information, a narrowly tuned ranker, or same-family verifier judgments.

Because runtime status can help verifier calibration, we test a stricter setting that masks runtime feedback during admission by setting $\mathrm{succ}(e_t)$ and $\mathrm{fail}(e_t)$ to unknown. Table~\ref{tab:no-runtime-feedback} shows that performance decreases modestly, but MemGuard remains above ReasoningBank in Table~\ref{tab:terminal-swe}. This confirms that the gains are not caused by consuming benchmark gold labels. In the default setting, runtime feedback refers only to information visible to the agent process, never hidden evaluator outcomes.

\begin{table*}[t]
\centering
\small
\setlength{\tabcolsep}{4pt}
\caption{Effect of masking runtime feedback during memory admission on Qwen-3.5-Plus.}
\label{tab:no-runtime-feedback}
\resizebox{\textwidth}{!}{%
\begin{tabular}{lcccccccc}
\toprule
\multirow{2}{*}{\textbf{Variant}}
& \multicolumn{2}{c}{\textbf{Terminal-Bench 2.0}}
& \multicolumn{2}{c}{\textbf{SWE-Bench Verified}}
& \multicolumn{2}{c}{\textbf{WebArena}}
& \multicolumn{2}{c}{\textbf{Mind2Web}} \\
\cmidrule(lr){2-3}\cmidrule(lr){4-5}\cmidrule(lr){6-7}\cmidrule(lr){8-9}
& SR & AS & Resolve Rate & AS & SR & AS & SSR & AS \\
\midrule
Default MemGuard & 67.4 & 30.4 & 83.6 & 30.6 & 58.4 & 6.9 & 51.8 & 11.5 \\
Mask runtime status & 66.6 & 31.0 & 82.4 & 31.3 & 56.5 & 7.2 & 50.3 & 11.9 \\
\bottomrule
\end{tabular}
}
\end{table*}

We also vary the retrieval and guard weights to check whether the hand-specified ranker is overly tuned. Table~\ref{tab:weight-sensitivity} compares the default development-stream weights with uniform weights, random Dirichlet weights averaged over five draws, and two simple verifier-aggregation alternatives on Qwen-3.5-Plus. The default is strongest, but uniform and random weights remain within a narrow band, suggesting that the result is not driven by a fragile weight setting.

\begin{table}[!htbp]
\centering
\small
\setlength{\tabcolsep}{2.5pt}
\caption{Weight sensitivity on Qwen-3.5-Plus. Random reports the mean over five Dirichlet weight draws.}
\label{tab:weight-sensitivity}
\resizebox{\columnwidth}{!}{%
\begin{tabular}{lcc}
\toprule
\textbf{Weights} & \textbf{WebArena SR} & \textbf{Mind2Web SSR} \\
\midrule
Default & 58.4 & 51.8 \\
Uniform retrieval/guard & 57.6 & 51.2 \\
Random retrieval/guard & 56.9 & 50.8 \\
Completion-heavy & 57.7 & 51.5 \\
Execution-heavy & 57.3 & 51.2 \\
\bottomrule
\end{tabular}
}
\end{table}

We also replace the verifier family in both directions to test whether the verifier signal simply echoes the agent backbone. Table~\ref{tab:cross-family-verifier} evaluates Qwen-3.5-Plus and Gemini-3.1-Pro agents with either same-family or cross-family verifiers across all four benchmarks. Cross-family verification slightly lowers absolute scores but preserves MemGuard's advantage over the matched ReasoningBank result in all eight agent--benchmark cells. This formal control shows that the central result does not depend on same-family self-evaluation alone.

\begin{table*}[t]
\centering
\small
\setlength{\tabcolsep}{3.2pt}
\renewcommand{\arraystretch}{1.12}
\caption{Cross-family verifier results across all four benchmarks. Each cell reports success metric / AS; matched ReasoningBank values appear in Table~\ref{tab:terminal-swe}.}
\label{tab:cross-family-verifier}
\begin{tabular*}{\textwidth}{@{\extracolsep{\fill}}llcccc}
\toprule
\textbf{Agent} & \textbf{Verifier} & \mbox{\textbf{Terminal-Bench 2.0}} & \textbf{SWE-Bench} & \textbf{WebArena} & \textbf{Mind2Web} \\
\midrule
\textbf{\mbox{Qwen-3.5-Plus}} & \mbox{Qwen-family} & 67.4 / 30.4 & 83.6 / 30.6 & 58.4 / 6.9 & 51.8 / 11.5 \\
\textbf{\mbox{Qwen-3.5-Plus}} & \mbox{Gemini-family} & 66.7 / 30.8 & 82.8 / 31.1 & 57.1 / 7.1 & 51.1 / 11.8 \\
\textbf{\mbox{Gemini-3.1-Pro}} & \mbox{Gemini-family} & 75.7 / 27.3 & 85.3 / 25.4 & 69.1 / 5.9 & 55.5 / 10.4 \\
\textbf{\mbox{Gemini-3.1-Pro}} & \mbox{Qwen-family} & 74.9 / 27.7 & 84.4 / 25.9 & 68.2 / 6.1 & 54.6 / 10.7 \\
\bottomrule
\end{tabular*}
\renewcommand{\arraystretch}{1.0}
\end{table*}

\subsection{Ablation Details}
\label{sec:ablation-details}

\begin{figure*}[t]
\centering
\includegraphics[width=0.92\textwidth]{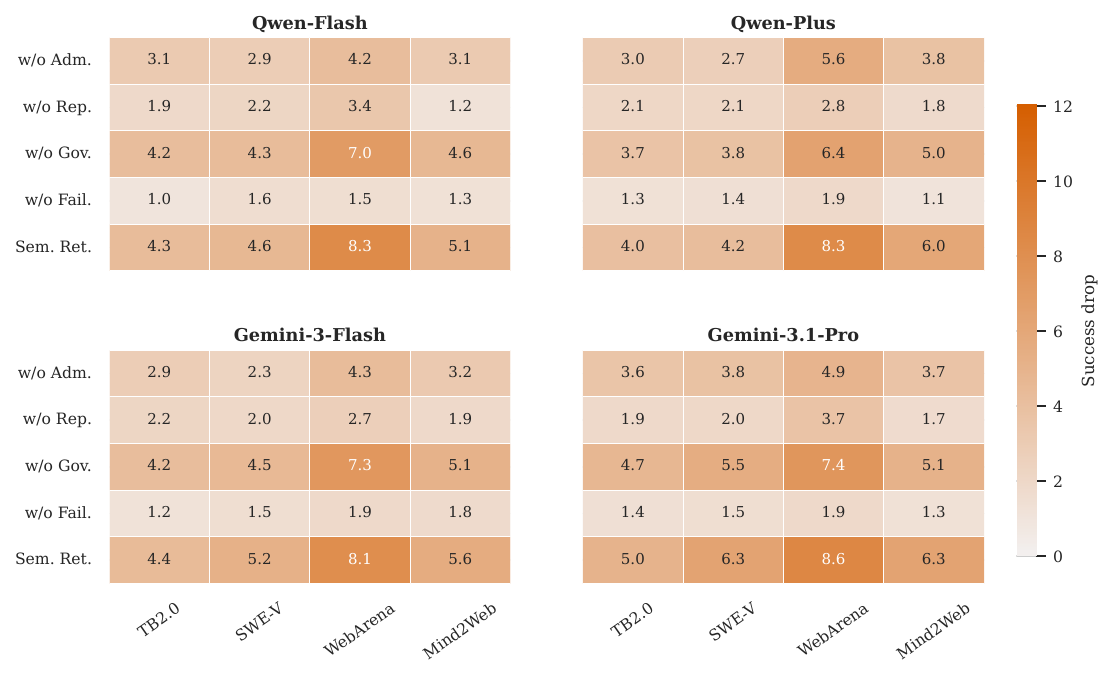}
\caption{Ablation impact across backbones. Each cell reports the success-metric drop from full MemGuard under a component removal or semantic-only retrieval control; darker cells indicate larger drops. The heatmap summarizes Table~\ref{tab:ablation}.}
\label{fig:ablation-heatmap}
\end{figure*}

Table~\ref{tab:ablation} reports the full ablation sweep across all four benchmarks and all four backbones. The compact main-text table in Table~\ref{tab:main-ablation} shows the Qwen-3.5-Plus slice, while this appendix table verifies that governance is the most important single-component removal for Qwen-3.5-Flash, Qwen-3.5-Plus, Gemini-3-Flash, and Gemini-3.1-Pro. Semantic-only retrieval performs worse, because it removes verifier- and conflict-aware ranking. Figure~\ref{fig:ablation-heatmap} visualizes the same results as success drops from the full system, making the relative importance of admission, governance, and retrieval scoring easier to inspect.

\begin{table*}[t]
\centering
\small
\setlength{\tabcolsep}{3.0pt}
\renewcommand{\arraystretch}{0.98}
\caption{Ablation results on all four backbones. Full denotes MemGuard with all components, and Sem. Ret. denotes semantic-only retrieval.}
\label{tab:ablation}
\begin{tabular}{lcccccccc}
\toprule
\multirow{2}{*}{\textbf{Variant}}
& \multicolumn{2}{c}{\shortstack{\textbf{Terminal-Bench}\\\textbf{2.0}}}
& \multicolumn{2}{c}{\shortstack{\textbf{SWE-Bench}\\\textbf{Verified}}}
& \multicolumn{2}{c}{\textbf{WebArena}}
& \multicolumn{2}{c}{\textbf{Mind2Web}} \\
\cmidrule(lr){2-3}\cmidrule(lr){4-5}\cmidrule(lr){6-7}\cmidrule(lr){8-9}
& SR $\uparrow$ & AS $\downarrow$
& \shortstack{Resolve\\Rate $\uparrow$} & AS $\downarrow$
& SR $\uparrow$ & AS $\downarrow$
& SSR $\uparrow$ & AS $\downarrow$ \\
\midrule
\rowcolor{modelrow}\multicolumn{9}{c}{\textbf{\textit{Qwen-3.5-Flash}}} \\
\rowcolor{bestrow}\textbf{Full} & \textbf{55.8} & \textbf{38.8} & \textbf{73.4} & \textbf{39.9} & \textbf{44.2} & \textbf{8.7} & \textbf{40.7} & \textbf{13.8} \\
w/o Adm. & 52.7 & 40.5 & 70.5 & 41.7 & 40.0 & 9.8 & 37.6 & 15.4 \\
w/o Rep. & 53.9 & 39.9 & 71.2 & 41.2 & 40.8 & 9.6 & 39.5 & 14.5 \\
w/o Gov. & 51.6 & 41.4 & 69.1 & 43.0 & 37.2 & 10.6 & 36.1 & 16.2 \\
w/o Fail. & 54.8 & 39.8 & 71.8 & 41.1 & 42.7 & 9.6 & 39.4 & 15.2 \\
Sem. Ret. & 51.5 & 41.5 & 68.8 & 43.1 & 35.9 & 11.0 & 35.6 & 16.7 \\
\midrule
\rowcolor{modelrow}\multicolumn{9}{c}{\textbf{\textit{Qwen-3.5-Plus}}} \\
\rowcolor{bestrow}\textbf{Full} & \textbf{67.4} & \textbf{30.4} & \textbf{83.6} & \textbf{30.6} & \textbf{58.4} & \textbf{6.9} & \textbf{51.8} & \textbf{11.5} \\
w/o Adm. & 64.4 & 32.0 & 80.9 & 32.5 & 52.8 & 8.0 & 48.0 & 13.1 \\
w/o Rep. & 65.3 & 31.5 & 81.5 & 31.7 & 55.6 & 7.4 & 50.0 & 12.2 \\
w/o Gov. & 63.7 & 32.6 & 79.8 & 33.0 & 52.0 & 8.5 & 46.8 & 13.8 \\
w/o Fail. & 66.1 & 31.4 & 82.2 & 31.5 & 56.5 & 7.8 & 50.7 & 12.8 \\
Sem. Ret. & 63.4 & 32.9 & 79.4 & 33.2 & 50.1 & 8.8 & 45.8 & 14.1 \\
\midrule
\rowcolor{modelrow}\multicolumn{9}{c}{\textbf{\textit{Gemini-3-Flash}}} \\
\rowcolor{bestrow}\textbf{Full} & \textbf{61.8} & \textbf{34.1} & \textbf{80.4} & \textbf{36.1} & \textbf{50.7} & \textbf{8.1} & \textbf{43.8} & \textbf{12.4} \\
w/o Adm. & 58.9 & 35.8 & 78.1 & 37.7 & 46.4 & 9.1 & 40.6 & 14.4 \\
w/o Rep. & 59.6 & 35.3 & 78.4 & 37.2 & 48.0 & 8.7 & 41.9 & 13.2 \\
w/o Gov. & 57.6 & 36.5 & 75.9 & 38.8 & 43.4 & 9.9 & 38.7 & 15.1 \\
w/o Fail. & 60.6 & 35.0 & 78.9 & 37.1 & 48.8 & 9.0 & 42.0 & 13.7 \\
Sem. Ret. & 57.4 & 36.6 & 75.2 & 39.0 & 42.6 & 10.3 & 38.2 & 15.6 \\
\midrule
\rowcolor{modelrow}\multicolumn{9}{c}{\textbf{\textit{Gemini-3.1-Pro}}} \\
\rowcolor{bestrow}\textbf{Full} & \textbf{75.7} & \textbf{27.3} & \textbf{85.3} & \textbf{25.4} & \textbf{69.1} & \textbf{5.9} & \textbf{55.5} & \textbf{10.4} \\
w/o Adm. & 72.1 & 29.1 & 81.5 & 27.0 & 64.2 & 6.7 & 51.8 & 11.1 \\
w/o Rep. & 73.8 & 28.2 & 83.3 & 26.4 & 65.4 & 6.5 & 53.8 & 11.2 \\
w/o Gov. & 71.0 & 29.4 & 79.8 & 27.9 & 61.7 & 7.2 & 50.4 & 12.7 \\
w/o Fail. & 74.3 & 28.0 & 83.8 & 26.3 & 67.2 & 6.4 & 54.2 & 11.4 \\
Sem. Ret. & 70.7 & 29.6 & 79.0 & 28.2 & 60.5 & 7.4 & 49.2 & 13.2 \\
\bottomrule
\end{tabular}
\end{table*}

\section{Implementation and Audit Protocols}
\label{sec:implementation-details}

This section specifies the implementation and audit details needed to reproduce the method-level behavior described in Section~\ref{sec:method}. Sections~\ref{sec:memory-schema}--\ref{sec:hyperparameters} define the stored record schema, verifier criteria, governance operations, and fixed hyperparameters. Sections~\ref{sec:diagnostic-protocols}--\ref{sec:memory-induction-example} document how the appendix diagnostics were audited and how a trajectory becomes a governed memory item. Experimental outcomes are reported separately in Appendices~\ref{sec:experiment-details}, \ref{sec:efficiency-ablation}, and~\ref{sec:robustness-governance}.

\subsection{Memory Record Schema}
\label{sec:memory-schema}

Each induced experience is stored as a structured record whose core fields include \textit{id}, \textit{type}, \textit{title}, \textit{description}, \textit{content}, \textit{applicability}, \textit{risk}, and \textit{evidence\_span}. Governance then attaches persistent metadata: \textit{quality\_score}, \textit{confidence}, \textit{verifier\_label}, \textit{source\_status}, lifecycle \textit{state}, \textit{usage\_count}, \textit{last\_used}, the \textit{deduplication\_signature}, \textit{conflict\_links}, and the verifier descriptor $d_m=(R_m,c_m,\ell_m,\nu_m)$. The \textit{state} field controls whether a memory is provisional, active, summarized, or archived. Retrieval uses only active and summary records by default, while failure-derived records can be rendered as constrained failure guards rather than positive action recipes.

\subsection{Verifier Criteria}
\label{sec:verifier-criteria}

The verifier evaluates each completed trajectory using task completion, evidence consistency, execution validity, and generalizability. Boundary cases are verified repeatedly with different views of the trajectory. MemGuard aggregates the resulting score-token distributions into trajectory reward, confidence, and a verifier label as described in Section~\ref{sec:method}. Low-reward or uncertain candidates are kept provisional or down-weighted during retrieval.

\subsection{Governance Operations}
\label{sec:governance-operations}

After admission, MemGuard compares new candidates with existing records to detect duplicate or conflicting experiences. Duplicate records are merged into compact summaries when they provide repeated evidence for the same reusable rule. Conflicting records are resolved by the lexicographic priority described after Equation~\ref{eq:conflict}. Weak, stale, or superseded records are archived when the active-memory budget is reached.

\paragraph{Reproduction checklist.}
For implementation-level reproducibility, we use a hybrid retriever that combines BM25 scores and embedding cosine similarity over the record title, description, content, applicability condition, and evidence span. Retrieval first forms a top-20 candidate pool, then injects up to five positive memories, two summary memories, and two failure guards after hard filtering. The deduplication signature concatenates memory type, abstract task pattern, normalized action category, tool or website scope, and key applicability condition. Merge and conflict checks use cosine similarity over the embedded record fields plus exact-match signature features; the merge threshold is 0.86 and the conflict threshold is 0.72. Governance runs after every completed task: activation, rejection, merge, and conflict checks are immediate, while summarization and archival are triggered whenever the active-memory budget is exceeded or a record becomes stale, low-quality, or superseded. Summaries are generated with the same memory-induction model using the prompt family in Appendix~\ref{sec:induction-prompt}; they preserve covered record ids, applicability conditions, and verifier metadata. Tie-breaking for conflicting records follows verifier label, reward, confidence, recency, and usage count.

\subsection{Hyperparameters and Implementation Details}
\label{sec:hyperparameters}

Table~\ref{tab:hyperparameters} lists the main hyperparameters used in all experiments. We use the same values across benchmarks unless a domain-specific memory budget is shown. Hyperparameters were selected once on a small held-out development stream by coarse grid search over thresholds and by setting retrieval weights to favor relevance, verifier quality, and conflict risk; they were then fixed for all reported runs. Table~\ref{tab:weight-sensitivity} compares these weights against uniform and random alternatives. The $\beta^+$ and $\beta^-$ rows follow the positive retrieval terms and risk terms in Equation~\ref{eq:positive-score}; the $\delta^+$ and $\delta^-$ rows follow the failure-guard score in Equation~\ref{eq:guard-score}. Weight orders are $\beta^+=(r,q,p,\mathrm{rec},\mathrm{use})$, $\beta^-=(\mathrm{cf},\mathrm{stale},\mathrm{ver})$, $\delta^+=(r,q,p)$, and $\delta^-=(\mathrm{cf},\mathrm{stale},\mathrm{ovr})$.

\begin{table}[!htbp]
\centering
\small
\setlength{\tabcolsep}{4pt}
\caption{Default MemGuard hyperparameters.}
\label{tab:hyperparameters}
\begin{tabular}{p{0.52\columnwidth}p{0.36\columnwidth}}
\toprule
\textbf{Parameter} & \textbf{Value} \\
\midrule
Criteria $C$ & 4 \\
Score tokens $\mathcal{V}_{\mathrm{score}}$ & $\{1,2,3,4,5\}$ \\
Token map $\phi(v)$ & $(v-1)/4$ \\
Default views $K_{\mathrm{view}}$ & 1 \\
Boundary views $K_{\mathrm{view}}$ & up to 3 \\
Uncertainty band & $[0.45,0.65]$ \\
Uncertainty mix $\alpha$ & 0.5 \\
Repeat threshold $\eta_{\mathrm{verify}}$ & 0.55 \\
Activation threshold & $R_t \geq 0.70$, $c_t \geq 0.60$ \\
Retrieved pool & 20 candidates \\
Injected positive memories & 5 records \\
Injected summaries & 2 records \\
Injected failure guards & 2 records \\
Failure guard relevance threshold $\eta_{\mathrm{guard}}$ & 0.62 \\
Failure guard confidence $\eta_{\mathrm{guard\_cnf}}$ & 0.70 \\
Web memory budget & 512 active records \\
Terminal/SWE budget & 384 active records \\
Merge threshold & 0.86 \\
Conflict threshold & 0.72 \\
Positive weights $\beta^+$ &
\begin{tabular}[t]{@{}l@{}}0.40, 0.25, 0.10,\\0.10, 0.15\end{tabular} \\
Risk weights $\beta^-$ & 0.20, 0.15, 0.30 \\
Guard positive weights $\delta^+$ & 0.45, 0.30, 0.25 \\
Guard risk weights $\delta^-$ & 0.15, 0.15, 0.25 \\
\bottomrule
\end{tabular}
\end{table}

The verifier prompt asks the model to score task completion, evidence consistency, execution validity, and generalizability using only the logged task, trajectory, final output, and runtime-visible execution status. Hidden benchmark labels and final evaluator outcomes are excluded. Each criterion must output a score token, a short rationale, and an evidence span. The full-trajectory view contains the complete trajectory; the evidence-focused view removes unrelated intermediate turns; and the risk-focused view asks the verifier to prioritize failed branches, ambiguous observations, and unsupported assumptions. Repeated verification uses deterministic decoding with criterion-order permutations rather than random sampling, which reduces variance while exposing disagreements across views.

\subsection{Diagnostic Analysis Protocols}
\label{sec:diagnostic-protocols}

The verifier calibration audit samples 100 completed trajectories from each benchmark under Qwen-3.5-Plus and Qwen-3.5-Flash. Within each benchmark-model slice, we stratify by verifier label when possible, targeting 40 \textit{verified\_success}, 32 \textit{verified\_fail}, and 28 \textit{uncertain} cases; if a slice contains too few cases of one label, the remainder is filled from the nearest confidence band. Two authors independently inspect the logged task, trajectory, final output, runtime-visible feedback, and benchmark feedback after evaluation, then mark whether the verifier's final label is supported by the available evidence. Across 800 audited cases, the two annotators agree on 724 cases, corresponding to Cohen's $\kappa=0.82$; disagreements are resolved by discussion before reporting Table~\ref{tab:verifier-calibration}. Benchmark feedback is used only for this offline audit, not for memory admission. A false accept means the verifier promotes an unsupported or failed trajectory as reusable; a false reject means it rejects or marks uncertain a trajectory that appears successful and transferable. We separately record whether a false accept would become active positive memory after governance. In 38 of 60 false-accept cases, the record remains provisional, is routed to a failure guard, or is blocked by conflict checks; the remaining 22 cases are counted as active false accepts and are traced in Table~\ref{tab:false-accept-tracing}.

For the noise-sensitivity study, we corrupt verifier labels before memory admission while leaving all other components fixed. Corruption randomly swaps labels among \textit{verified\_success}, \textit{verified\_fail}, and \textit{uncertain}; reward and confidence are adjusted consistently by moving corrupted records toward the nearest threshold interval. This setting is intentionally harsher than ordinary verifier variance because it can promote bad candidates and suppress good ones.

The failure-memory audit samples 200 failure-derived candidate memories whose source trajectory or later retrieval context exposes a plausible risk, stratified across WebArena and Mind2Web and across the five risk types in Table~\ref{tab:failure-risk}. Two annotators independently classify the primary risk type and governance outcome; agreement is 177/200 with Cohen's $\kappa=0.86$. Disagreements are resolved by discussion. We classify the primary risk type by reading the source trajectory, induced memory, verifier rationale, and later retrieval context. ``Active'' means that the record remains retrievable as a constrained failure guard after governance, whereas ``governed'' means that the record is rejected, kept provisional, archived, or merged into a safer summary before it can be used as an action suggestion.

\subsection{Memory Induction Example}
\label{sec:memory-induction-example}

The memory induction module is prompt-based; the exact template is shown in Appendix~\ref{sec:induction-prompt}. Given a verifier-processed trajectory, it first identifies evidence spans that explain success or failure. It then rewrites instance-specific spans into reusable conditions by replacing concrete paths, DOM identifiers, issue names, and command arguments with abstract task patterns. Finally, it emits records using fixed core fields: \textit{type}, \textit{title}, \textit{description}, \textit{content}, \textit{applicability}, \textit{risk}, and \textit{evidence\_span}; governance then attaches verifier metadata, lifecycle state, and conflict links. Candidates without evidence spans are rejected before activation. This schema prevents the system from storing unsupported reflections and makes each memory auditable.

\begin{figure}[t]
\centering
\includegraphics[page=1, trim=223bp 470bp 71bp 54bp, clip, width=\columnwidth]{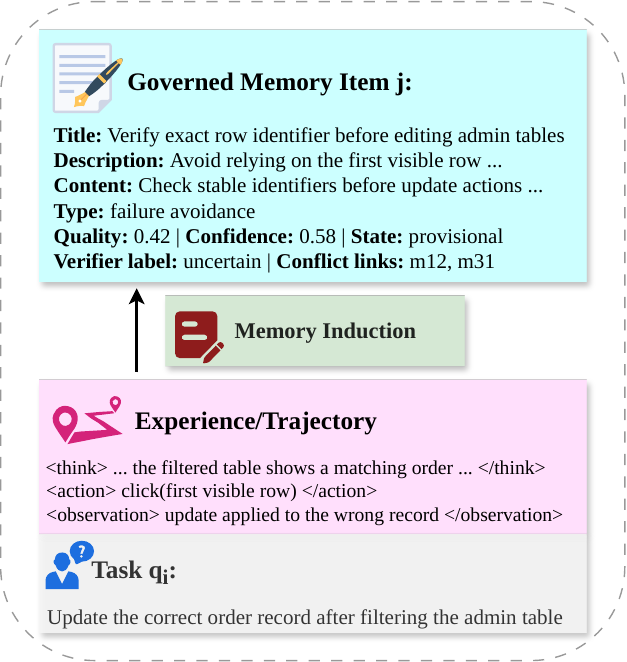}
\caption{Example of memory induction. A trajectory with an unsupported table-edit action is distilled into a governed memory item with content fields and persistent metadata, including type, quality, confidence, state, verifier label, and conflict links. The uncertain, low-confidence item remains a failure-avoidance guard rather than a positive action recipe.}
\label{fig:memory-item-example}
\end{figure}

Figure~\ref{fig:memory-item-example} illustrates this process with a web-admin failure. The logged trajectory contains an incorrect action that edits the first visible row after filtering; the induced memory is therefore represented as a failure-avoidance record with provisional state, low confidence, and an uncertain verifier label rather than as a reusable action recipe.

\section{Robustness and Governance Analyses}
\label{sec:robustness-governance}

This section focuses on whether governed memory remains stable under memory pressure and failure-derived records. It reports memory-bank health, memory-budget sensitivity, failure-memory risk analysis, a qualitative case study, and paired bootstrap significance tests. These diagnostics complement the main success metrics by showing how the memory bank evolves and how reported gains are checked statistically.

\subsection{Memory Health and Governance Analysis}
\label{sec:governance-analysis}

Task success alone does not show whether a memory system remains healthy over a long task stream. We therefore track governance-level statistics that directly correspond to MemGuard's design: the final number of active memories, the fraction of candidate memories rejected before activation, the fraction merged into existing records, the fraction archived by governance, and the fraction marked uncertain by repeated verification. These statistics measure whether the memory bank is merely growing or actively maintaining a compact and reliable active set.

\begin{table}[t]
\centering
\footnotesize
\setlength{\tabcolsep}{1.6pt}
\caption{Memory health statistics on Qwen-3.5-Plus. Act. is the final active-memory count; the remaining columns are percentages of induced candidate memories.}
\label{tab:memory-health}
\begin{tabular}{@{}lrrrrrr@{}}
\toprule
\textbf{Benchmark} & Act. & Rej. & Merge & Arch. & Unc. & Conf. \\
\midrule
Terminal-Bench 2.0 & 286 & 18.4 & 22.7 & 14.1 & 9.6 & 6.8 \\
SWE-Bench Verified & 342 & 21.3 & 19.5 & 16.8 & 11.2 & 8.4 \\
WebArena & 418 & 29.7 & 24.1 & 21.5 & 17.9 & 13.6 \\
Mind2Web & 463 & 32.5 & 26.8 & 24.2 & 19.4 & 15.1 \\
\bottomrule
\end{tabular}
\end{table}

\begin{figure*}[!t]
\centering
\includegraphics[width=0.98\textwidth]{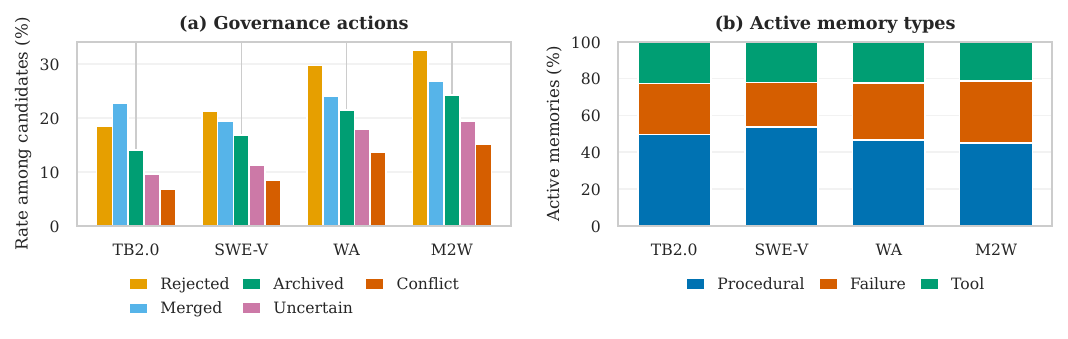}
\vspace{-0.5em}
\caption{Memory-governance diagnostics on Qwen-3.5-Plus. Left: non-mutually-exclusive governance action rates over induced candidate memories; a candidate can trigger multiple actions or none of these actions, so the bars need not sum to 100\%. Right: active-memory type composition after governance. The figure visualizes Tables~\ref{tab:memory-health} and~\ref{tab:memory-types}.}
\label{fig:memory-governance}
\vspace{-0.6em}
\end{figure*}

Table~\ref{tab:memory-health} and Figure~\ref{fig:memory-governance} show that web benchmarks produce more rejected, uncertain, and conflict-resolved candidates than terminal or software engineering benchmarks. These columns are not mutually exclusive counts, so they should be read as governance diagnostics rather than a partition of all candidates. This pattern is consistent with the nature of web interaction: small changes in page state, website policy, or navigation history can make an experience misleading even when its surface description appears relevant. MemGuard does not treat these candidates as harmless noise. Instead, it keeps uncertain records out of active memory, merges repeated evidence into compact summaries, and archives weak or conflicting entries. This analysis supports the paper's central claim that reliable long-term experience reuse requires governance signals beyond semantic relevance.

We also inspect the composition of active memory types in Table~\ref{tab:memory-types}. Procedural hints remain the largest category across benchmarks, supporting the experience-reuse framing. Web benchmarks nevertheless contain a larger share of failure-avoidance memories because incorrect clicks, unstable page states, and misleading table orders frequently produce negative lessons. Tool-usage memories are comparatively stable across domains. This analysis explains why removing failure memories mainly affects AS: these memories often prevent repeated exploration even when they do not directly solve the task.

\begin{table}[t]
\centering
\small
\setlength{\tabcolsep}{4pt}
\caption{Active memory type distribution on Qwen-3.5-Plus. Values are percentages of active memories after governance.}
\label{tab:memory-types}
\begin{tabular}{lccc}
\toprule
\textbf{Benchmark} & \textbf{Proc.} & \textbf{Fail.} & \textbf{Tool} \\
\midrule
Terminal-Bench 2.0 & 49.7 & 27.6 & 22.7 \\
SWE-Bench Verified & 53.8 & 24.1 & 22.1 \\
WebArena & 46.8 & 30.8 & 22.4 \\
Mind2Web & 45.1 & 33.5 & 21.4 \\
\bottomrule
\end{tabular}
\end{table}

\subsection{Memory Budget Sensitivity}
\label{sec:budget-sensitivity}

\begin{figure*}[!t]
\centering
\includegraphics[width=0.98\textwidth]{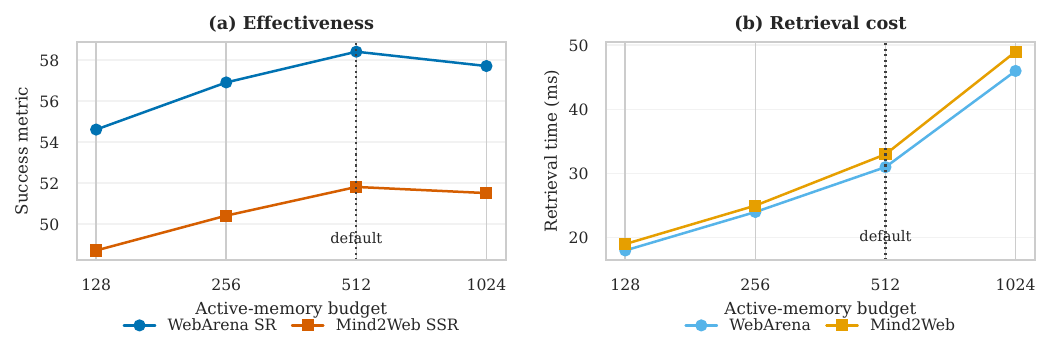}
\vspace{-0.5em}
\caption{Memory-budget sensitivity on Qwen-3.5-Plus. Left: task success metrics under different active-memory budgets. Right: retrieval latency before prompt construction.}
\label{fig:budget-sensitivity}
\vspace{-0.6em}
\end{figure*}

A long-running memory system should not rely on unbounded growth. We therefore vary the active-memory budget while keeping the retrieval budget fixed. Table~\ref{tab:budget-sensitivity} and Figure~\ref{fig:budget-sensitivity} report Qwen-3.5-Plus on WebArena and Mind2Web, the two settings where memory pressure is highest. Very small budgets discard useful domain-specific patterns; very large budgets retain more marginal and conflicting memories, increasing retrieval latency and slightly reducing precision. The default budget is near the best trade-off between effectiveness and memory overhead.

\begin{table}[!htbp]
\centering
\small
\setlength{\tabcolsep}{4pt}
\caption{Memory-budget sensitivity on Qwen-3.5-Plus. Retrieval time is measured before prompt construction.}
\label{tab:budget-sensitivity}
\begin{tabular}{lcccc}
\toprule
\multirow{2}{*}{\textbf{Budget}}
& \multicolumn{2}{c}{\textbf{WebArena}}
& \multicolumn{2}{c}{\textbf{Mind2Web}} \\
\cmidrule(lr){2-3}\cmidrule(lr){4-5}
& SR & Ret. ms & SSR & Ret. ms \\
\midrule
128 & 54.6 & 18 & 48.7 & 19 \\
256 & 56.9 & 24 & 50.4 & 25 \\
\textbf{512} & \textbf{58.4} & \textbf{31} & \textbf{51.8} & \textbf{33} \\
1024 & 57.7 & 46 & 51.5 & 49 \\
\bottomrule
\end{tabular}
\end{table}

\subsection{Failure-Memory Risk Analysis}
\label{sec:failure-risk}

Learning from failed trajectories is useful only if the system prevents failure-derived memories from becoming misleading action recipes. We audit 200 failure-derived candidate memories from WebArena and Mind2Web for which the source trajectory or later retrieval context exposed a plausible risk. Table~\ref{tab:failure-risk} categorizes the main risk in each case and records whether MemGuard allowed the item to remain active as a constrained failure guard or governed it through rejection, provisional retention, merging, or archival. The most common risks are over-generalized guards and stale page-state assumptions; both are expected in web tasks because surface descriptions can remain similar while the correct action depends on a hidden identifier, sort order, or page state.

\begin{table}[!htbp]
\centering
\scriptsize
\setlength{\tabcolsep}{3pt}
\caption{Audit of failure-memory risks across 200 WebArena and Mind2Web cases. Counts are cases where the listed risk was the primary issue; Active means the item remains retrievable as a constrained failure guard, and Governed means it is rejected, kept provisional, merged, or archived before use as an action suggestion. Two-annotator agreement is $\kappa=0.86$.}
\label{tab:failure-risk}
\begin{tabular}{p{0.43\columnwidth}ccc}
\toprule
\textbf{Risk type} & Count & Active & Governed \\
\midrule
Incorrect strategy extraction & 36 & 5 & 31 \\
Misattributed failure cause & 44 & 8 & 36 \\
Over-generalized guard & 56 & 12 & 44 \\
Over-specific guard & 28 & 4 & 24 \\
Stale or conflicting state & 36 & 2 & 34 \\
\bottomrule
\end{tabular}
\end{table}

The audit supports the separation between positive recall and negative guards. Only 31 risky records become active, and those records are injected as constraints rather than as direct action plans. The remaining cases are rejected, kept provisional, merged into a safer summary, or archived after conflict detection. Appendix~\ref{sec:diagnostic-protocols} gives the annotation protocol. This behavior explains why removing failure memories in Table~\ref{tab:ablation} mainly increases AS: useful failure guards reduce wasted exploration, while the governance layer limits their ability to override the current observation.

\subsection{Case Study: Avoiding Misleading Experience}
\label{sec:case-study}

We further inspect a representative WebArena task in which the agent must update an order record in an admin interface after applying a specific filter. A previous failed trajectory produced the memory: ``When the order table is visible, edit the first returned row after searching by customer name.'' This memory is superficially relevant to later admin tasks, but the earlier failure occurred because the table was sorted by update time rather than by exact search match. A ReasoningBank-style baseline retrieves this high-level experience as useful guidance and edits the first row, causing the later task to fail.

MemGuard handles this case differently. During admission, the verifier assigns low evidence consistency because the earlier action is not supported by a stable page observation, and low generalizability because the rule depends on an unstable table ordering. Repeated verification disagrees on task completion, so the candidate is labeled \textit{uncertain} with a low reward. MemGuard therefore keeps the memory provisional rather than active; if the record is retained for later audit, its verifier penalty lowers its retrieval weight. When a similar task appears later, the unstable rule is not injected; instead, the retrieved memory advises the agent to verify the exact row identifier before editing. This example illustrates how verifier reward and uncertainty labels prevent a relevant but wrong experience from becoming a persistent source of errors.

\subsection{Significance Tests}
\label{sec:significance-tests}

We compute the main significance test with paired nonparametric bootstrap resampling over benchmark tasks. For each backbone--benchmark setting, we first average the paired MemGuard-minus-ReasoningBank outcome over the five matched runs, then resample task IDs with replacement 10{,}000 times while retaining method pairing. This treats tasks, rather than task-seed observations, as the bootstrap unit. Table~\ref{tab:bootstrap-pvalues} reports the resulting raw $p$-values.

\begin{table}[!htbp]
\centering
\footnotesize
\setlength{\tabcolsep}{3pt}
\caption{Raw paired-bootstrap $p$-values for MemGuard vs. ReasoningBank on the primary success metric. TB2.0 and WA use SR, SWE-V uses Resolve Rate, and M2W uses SSR.}
\label{tab:bootstrap-pvalues}
\resizebox{\columnwidth}{!}{%
\begin{tabular}{lcccc}
\toprule
\textbf{Backbone} & \textbf{TB2.0} & \textbf{SWE-V} & \textbf{WA} & \textbf{M2W} \\
\midrule
Qwen-3.5-Flash & 0.018 & 0.011 & $<0.001$ & 0.006 \\
Qwen-3.5-Plus & 0.021 & 0.013 & $<0.001$ & $<0.001$ \\
Gemini-3-Flash & 0.026 & 0.019 & 0.002 & 0.004 \\
Gemini-3.1-Pro & 0.032 & 0.017 & $<0.001$ & 0.002 \\
\bottomrule
\end{tabular}
}
\end{table}

Applying Benjamini--Hochberg correction over these 16 prespecified tests gives $q=0.004$--$0.032$, so all 16 remain significant at FDR 0.05. Under Holm family-wise correction, conservatively treating each printed $p<0.001$ as $p=0.001$, 7/16 remain significant: all four WebArena comparisons and the Mind2Web comparisons for Qwen-3.5-Plus, Gemini-3-Flash, and Gemini-3.1-Pro. The five-seed means therefore all favor MemGuard, but individual runs are less uniform. Across the 320 matched setting--seed--metric comparisons against ReasoningBank and Verifier-only Filter (16 settings $\times$ 5 seeds $\times$ 2 metrics $\times$ 2 baselines), MemGuard has 11 losses and one tie at the reported precision.

\FloatBarrier
\section{Reproducibility Statement}
\label{sec:reproducibility}

All methods are evaluated with matched step budget, decoding settings, retrieval budget, and memory-context budget. AS is reported as average agent action steps per task, while verifier calls, token counts, and storage overhead are reported separately. We specify the memory schema, prompts, model versions, verifier settings, retrieval budgets, governance thresholds, and table/figure generation procedure in the appendix.

\section{Use of LLMs}
\label{sec:llm-use}

LLMs are used as backbone agents, verifiers, and memory-induction modules in the experiments described in this paper. The paper draft was edited with AI assistance for language polishing, organization, and LaTeX formatting. All scientific claims, experiment design choices, benchmark descriptions, and reported numbers remain the authors' responsibility.
\end{document}